\documentclass{article}
\usepackage{cite}
\usepackage{amsmath,amssymb,amsfonts}
\usepackage{algorithmic}
\usepackage{graphicx}
\usepackage{textcomp}
\usepackage{array}
\usepackage{url}
\usepackage{listings}
\usepackage[normalem]{ulem}

\title{Virtual Experience to Real World Application: Sidewalk Obstacle Avoidance Using Reinforcement Learning for Visually Impaired}

\author{Faruk Ahmed, Md Sultan Mahmud \\Kazi Ashraf Moinuddin, Mohammed Istiaque Hyder \\ and Mohammed Yeasin \\	mfahmed@memphis.edu; mmahmud@memphis.edu\\ kmnuddin@memphis.edu; mohammedhyder121@gmail.com\\ myeasin@memphis.edu}

\date{}

\begin{document}

\maketitle

\begin{abstract}
Finding a path free from obstacles that poses minimal risk is critical for safe navigation. People who are sighted and people who are visually impaired require navigation safety while walking on a sidewalk. In this research we developed an assistive navigation on a sidewalk by integrating sensory inputs using reinforcement learning. We trained a Sidewalk Obstacle Avoidance Agent (SOAA) through reinforcement learning in a simulated robotic environment. A Sidewalk Obstacle Conversational Agent (SOCA) is built by training a natural language conversation agent with real conversation data. The SOAA along with SOCA was integrated in a prototype device called augmented guide (AG). Empirical analysis showed that this prototype improved the obstacle avoidance experience about 5\% from a base case of 81.29\%.  

\end{abstract}

\section{Introduction}

\begin{figure}[!h]
	\begin{center}
		\includegraphics[width=0.8\linewidth]{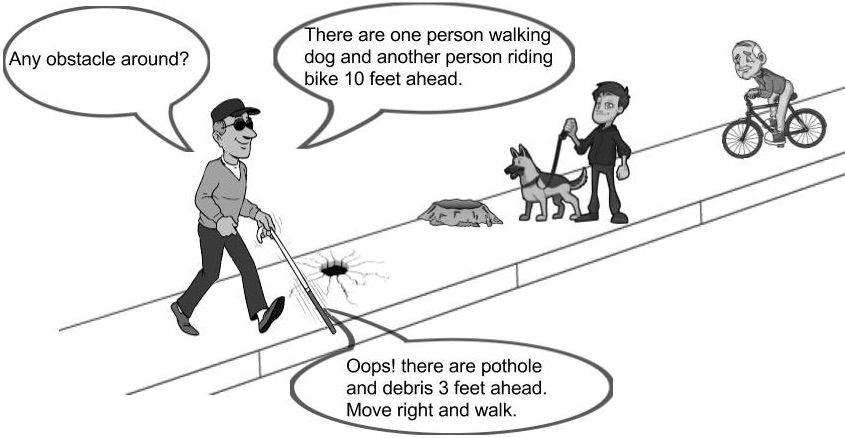}
    \end{center}
	\caption{The blind person walking on the sidewalk with the help of AG.}
	\label{fig:ambienttheme}
\end{figure}

According to the estimates from the World Health Organization (WHO) about 285 million people are visually impaired worldwide: 39 million are blind, and 246 million have low vision \cite{bourne2017magnitude,fricke2018global}. 2.3\% people have a visual disability in USA according to the Disability Status Report \cite{erickson2008disability}. In addition, according to traffic safety facts data, there were 4,735 pedestrians killed in 2013 in United States \cite{national2017traffic}.  Hence, ambient awareness on the sidewalk is critical for safe navigation for the people who are blind or visually impaired.  Ambient awareness may include (but not limited to):  potholes, debris, ongoing construction, as well as a person riding a bike or walking a pet. Fig. \ref{fig:ambienttheme} depicts a typical scenario of ambient awareness on a sidewalk.  Despite the needs and advances in assistive technologies, designing a functional and easy to use system remains a challenge. 

There are copious number of electronic travel aids (ETAs) to help blind people with obstacle detection, obstacle avoidance, and navigation. Blind people often use cane, smart phone apps or other ETA devices while walking. These technologies provide some information for safe navigation but are not successful in many cases. For example, global positioning system (GPS) devices provide direction only but do not notify about subtle obstacle e.g., potholes, puddles, or traffic cones. Visually impaired people use guide cane to become aware of paths and obstacles by continuously striking on the surface. Slippery surfaces and slope are not detected by guide canes. There are some mobile apps that provide turn-by-turn instructions, but do not provide ambient awareness. A few applications detect obstacles in indoor and outdoor environments except identifying sidewalk obstacles \cite{aladren2016navigation}. Most of the reported applications are not interactive. Those applications are limited to single mode operations (e.g., audio feedback or haptics). In addition, the design of feedback mechanisms suffer from insufficient personalization and reconfigurability. This induces more cognitive load. To reduce the cognitive load, there is a research to apply image captioning for sidewalk ambient awareness \cite{ahmed2018image}.

While walking on a sidewalk, a visually impaired person needs enough information to create a mental map of the environment. A meaningful description (e.g., ``a person with bike is coming towards you'') of a visual scene that could be useful for assistive solutions. The incomplete description may lead to poor perception and misrepresentation of dangers that may lie ahead. Incomplete or partial characterization of a scene such as ``person with bike'' can be irrelevant or meaningless feedback to a person who is blind. The first sentence is an example of a caption or description of an image, whereas the second sentence is the example of a class label. On the other hand, meaningful feedback should not be too long and also should not induce excessive cognitive load. Hence, ``meaningful feedback'' is crucial for safer navigation.

Recently the Deep Neural Network (DNN) solved some problems efficiently. Researchers from both academia and industries have been using the power of DNNs for speech recognition \cite{abdel2014convolutional}, text categorization \cite{wang2012end} and image recognition \cite{deng2009imagenet}, just to name a few. Convolutional Neural Networks (CNN)-based deep learning architectures are state-of-the-art for visual recognition tasks. Reinforcement Learning (RL) is able to beat human playing game  \cite{bellemare2013arcade,mnih2015human}. Recurrent Neural Network (RNN) is generating sentences and recognizing spoken language \cite{mikolov2010recurrent,graves2013speech}. In this research, we attempt to bridge the gap between DNN and assistive technology solutions. DNN-based assistive technology solutions can be useful to augment the natural sense-ability of the visually impaired, and this can be an aid for them.

\section{Context}
Assistive technology solutions for the visually impaired drew attention of researchers as a prominent research area in the mid-90s. Researchers conduct studies and develop applications to improve the mobility of visually impaired. Generally two types of applications are available for visually impaired \textit{a) standalone device b) mobile apps}. Classical computer vision and DNN are main technology for image-based assistive solution. 

\textbf{Classical computer vision applications:} Rao et al. \cite{rao2012acoustic} used the video and frame by frame processing to detect a pre-modeled obstacle and provides three different pitches of sound to avoid obstacles. On the other hand, Aladren et al. \cite{aladren2016navigation} used RGB-D sensor and computer vision technique to segment the floor in the indoor environment to find the barrier. However, this system is cumbersome and does not address the issues in a dynamic outdoor environment (sidewalk). Leung et al. \cite{leung2014visual} developed the application which determines the egomotion in the highly dynamic outdoor environment. The purpose of this application is to predict the visual odometry to help navigate blind people, based on predefined map, it does not provide ambiance information. However, none of these applications is capable of catering ambient awareness. 

\textbf{Deep learning applicatios:} Google Goggles is used to ``search'', based on pictures taken by handheld devices. The Orcam MyEye \cite{orcam} is efficient in reading text, road-sign, and traffic signal. ``Clarifai'' developed the image recognition engine whose underlying technique is Deep Convolutional Neural Networks (DCNN). Beside those commercial apps, Szegedy et al. \cite{deepnetobstacle}  used DNN to detect and localize objects of various classes including bicycle, dog, person, car and bus. 
SqueezeNet achieves AlexNet-level accuracy on ImageNet with fifty times fewer parameters \cite{iandola2016squeezenet}. This is a smaller network with comparable performance. Another important network announced by Google is the MobileNet \cite{howard2017mobilenets}. This network is built to achieve a balanced trade-off between accuracy and resources available in mobile hand-held device. DNN require huge computing resource to train and test as well as in production. That is why optimizing DNN became an important branch of the research. Research work related to optimizing the performance of DNN the deep compression network \cite{denton2014exploiting} is mentionable. Various methods based on vector quantization \cite{gong2014compressing}, hashing techniques \cite{chen2015compressing}, circulant projection \cite{cheng2015fast}, and tensor train decomposition \cite{novikov2015tensorizing} were reported with better compression capability.

In \cite{shinohara2016designing} Shinohara $et~al.$ performed a study to investigate the designers regard disability and accessible design thinking for disabled and non-disabled population. According to them, designing for both population simultaneously surface challenges and tensions lead better accessible design. Kawas $et~al.$ in \cite{kawas2016improving} performed qualitative study to understand real-time captioning experiences of deaf and hard of hearing (DHH) students in classroom setup. Their discovery is that the accuracy and reliability of the technology are still the most important issue of current captioning solutions. Wilson $et~al.$ in \cite{wilson2015using} present a study suggesting the accuracy of peripersonal reaching can be improved by the use of dynamic sound from both the objects to reach for and the reaching hand itself (via a word speaker) that changes based on the proximity of the hand to the object. Part of this research is useful for the ambient awareness application on a sidewalk because if an assistive technology solution produces dynamic sound, it will help the blind person to draw a mental map of the ambiance. Kane $et~al.$ in \cite{kane2009freedom} interviewed 20 participants with visual and motor disabilities and asked about their current use of mobile devices, including how they select them, how they use them while away from home, and how they adapt to accessibility challenges when on the go. They show that people with visual and motor disabilities use a variety of strategies to adapt inaccessible mobile devices and use them to perform everyday tasks and navigate. The assistive solution with accessible design will help visually impaired improve their daily life, and they will be able to carry out everyday tasks with ease.

Among the standalone devices Drishti \cite{helal2001drishti} and GuideCane \cite{ulrich2001guidecane} used GIS information hosted on a central server. They continuously queried the server for GPS information to facilitate navigation. GuideCane used ultrasonic sensor and embedded computer to detect obstacles, but the field of view of the sensor is very narrow. To circumvent the problem, Shoval et al. proposed an array of ultrasonic sensors  mounted on a belt \cite{shoval2003navbelt}. However, it became too bulky along with power and resource hungry. To make it lighter, the smart cane  project focused on obstacles that are of head height \cite{wu2008smartcane,singh2010smart,wahab2011smart}. A talking navigation cane was reported in \cite{jesie2015advanced}, that allows voice command and provide navigation information via audible messages and haptic feedback. They used the GPS to accommodate the localization of the person. With the revolution of the smart phone, the focus shifted towards developing the vision-based systems as well as assistive apps. These systems can be broadly categorized into perceiving by computer-vision and machine-learning algorithms.

\section{Problem}
``One guide dog takes about two years to train and costs a total of \$45,000 to \$60,000, covering everything from boarding a dog to extensive drilling by professional trainers in serving the needs of the blind to a weekslong period acclimating dog to recipient.'' \cite{sullivan_2013} The central question is can we do better? can we have an alternative to train a machine at an affordable cost? If so who do we train as a guide? This work is an attempt to answer the question. 

\subsection{Challenges}
Researchers build systems integrating deep architectures, GPS, as well as image captioning. But the problem of training someone as a guide is still unsolved because those intelligent system does not really understand the need of a visually impaired. It is like, ``here is a system go and use it'' instead of ``here is a system you can train yourself and use it''.

There are numerous challenges. 
\begin{itemize}
    \item System design: Should it be hand held or wearable? If it is hand held then it has to be light weight without compromising power supply. On the other hand if it is wearable, how easy it would be to wear? Glass on the eye, battery in pocket, cell phone in another pocket, does not seem like a very user friendly. 
    \item Network connectivity: For making the device intelligent, there is need pf computing resources which is not possible to carry as of today. Alternative is to use the data network. Keep in mind that data does not come free. Using wifi or cellular data is another challenge to incorporate. 
    \item Power consumption: The device must be low power consuming. 
    \item Efficient feedback: The feedback must be meaningful to the user. Without adding much cognitive load. Not continuous talking but required amount of talk. 
    \item Improvement from failure: There has to a way for the device to learn from failure. Usage analytic come handy in this case.
\end{itemize}

\section{System design}

To design an assistive system for the visually it is important to incorporate design thinking, system thinking and assistive thinking. We have worked with visually impaired keeping in mind these three design philosophies. 

Design thinking is a ``complex processes of inquiry and learning'' that designers carry out in a systems development process, ``making decisions as they proceed'' by ``working collaboratively on teams'' \cite{dym2005engineering}.

System thinking focuses on recognizing the interconnections between the parts of a system and then synthesizes them into a uniﬁed view of the whole \cite{assaraf2005development,kim1994putting}.

Assistive thinking is a hybrid approach ``to identify user's preference and resources early'' and ``incorporate them implementing the design'' that addresses a ``specific disability in the best possible manner'' \cite{hossain2013assistive}. 

In few sentences the system takes input from the environment, sense obstacle, and provide feedback to the user. In this research is about the evolution of feeding input, sensing obstacle and providing feedback.  

\subsection{Design thinking}

We started with identifying most important sidewalk obstacle from a visually impaired user's perspective. A pilot study provided a list of obstacles. 

\subsubsection{Identify important sidewalk obstacles}
\begin{figure}[h]
	\begin{center}
		\includegraphics[scale=0.6]{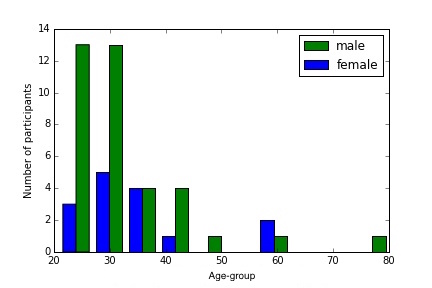}
	\end{center}
	\caption{Distribution of participants.}
	\label{fig:participants}
\end{figure}

To select relevant obstacles on the sidewalk, we interviewed $50$ visually impaired people \footnote{IRB approvals at University of Memphis 16322937, 22627312, 29904158}. Among the participants, 20 people have complete vision loss (10 of them are congenitally blind), and 30 have partial sight of different degrees. We conducted the interviews by asking them open-ended questions. The questions were: (1) how often do you walk on a sidewalk? (2) what are the difficulties you face on a sidewalk? (3) how do you resolve those difficulties? From their answers the top 10 difficulties (obstacles) are listed in Table \ref{tbl:listobstacle}. The distribution of the age group and gender group of the participants is presented in the Figure \ref{fig:participants}. From the list of obstacles, we observe that some obstacles seemed less important (e.g., crowd) but showed up in the list with 18\%. The crowd is visible more in urban areas compared to rural areas. This implies that the list of obstacles is a general list not area specific. 

\begin{table}[h]
	\caption{List of top 10 obstacles identified by representative users.}
	\label{tbl:listobstacle}
    \setlength{\tabcolsep}{3pt}
    \centering
	\begin{tabular}{|c|c|}
		\hline
		Name & Percentage (\%) \\
		\hline
		Potholes / Damaged Sidewalk & 29 \\
		\hline
		Crowd & 18 \\
		\hline
		Construction & 14 \\
		\hline
		Person with pet & 8 \\
		\hline
		Person with bike / bike & 6 \\
		\hline
		Curbs & 4 \\
		\hline
		Slope & 4 \\
		\hline
		Poles & 3 \\
		\hline
		No sidewalk (sidewalk ending) & 3 \\
		\hline
		Narrow sidewalk & 2 \\
		\hline
		
	\end{tabular}
\end{table}

Once we identified the obstacles. The users were interviewed and mainly they wanted to have some kind of feedback about the obstacle to avoid it. Visually impaired people outperform sighted people in determining direction, distance of source of a sound \cite{nilsson2016blind}. This is because the brain reorganizes the tasks of visual cortex along with auditory cortex due to lack of visual sensory information \cite{merabet2010neural}. That is why we proceeded with simple image classification task to play out the identified obstacle's name as feedback.  

\subsubsection{Obstacle image dataset}
Image classification requires databases to train a deep neural network (e.g., convolutional neural network (CNN)). There are widely used image databases available, e.g., MNIST, ImageNet, CIFAR, Caltech, STL-10 \cite{coates2011analysis}, SVHN, NIST. These databases cover many classes of objects from handwritten digits to house numbers; vehicles to animals. Manduchi et al. did a thorough study on mobile vision \cite{manduchi2012mobile}. They also created an inertial sensor time series dataset which can be used to model turn taking, step counting of blind people \cite{flores2018weallwalk}. However, there is no custom database related to the obstacles on a sidewalk. Most existing databases do not have objects affected by diurnal cycles and shadows. Therefore, we decided to create AS image database that incorporates obstacles identified by representative users. To the best of our knowledge, there is no publicly available database that specifically collects for sidewalk obstacles. 

\begin{figure}[bth]
	\centering
	\includegraphics[width=\linewidth]{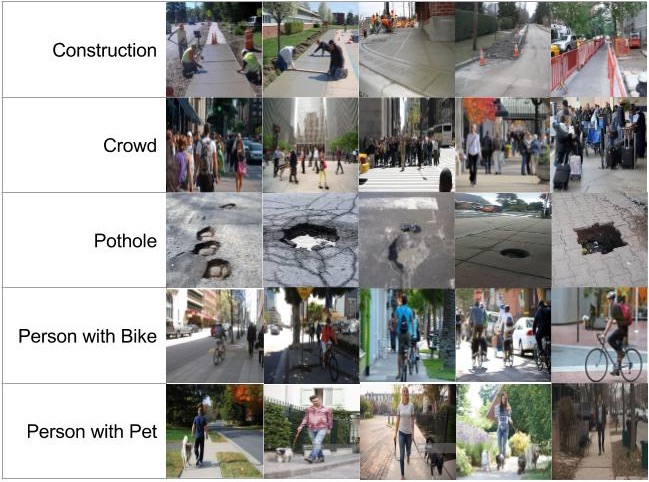}
	\caption{Samples from the sidewalk obstacle dataset.}
	\label{fig:dataset}
\end{figure}

In this pilot dataset, we consider five classes of obstacles: construction, crowd, pothole, person with bike, and person with pet. Each class has $10,000$ images, and the total numbers of images are $50,000$. Each image is $96\times96$ pixels with RGB channels. The size of the images was inspired by STL-10  natural image dataset \cite{coates2011analysis}. Figure \ref{fig:dataset} shows the sample data from the database \cite{sidewalkdataset}. The images were collected by taking photographs from the sidewalk and also using Google image search. We also used virtual example creation to increase the size of the database for training and testing the models.

There are noticeable variabilities in the image database. 25\% of the images have occlusion. Some of them have multiple obstacles, for example, construction and person-pet, construction and person-bike, pothole and construction. The database contains blurred images, images captured under diurnal cycle from very low light to bright light, affine transformed images that may occur due to the position of the sensor on the obstacles, different types noises to account for ambient conditions. These variabilities make image recognition difficult. Some of the images contain shadow.
Seasonal variations such as snow or rainfall are not included in this version of the database. In addition, it does not have any image of different terrain as well.

Again we spoke to the visually impaired people. The name of the obstacle was too little information. They wanted to know more about the obstacle. For this reason, we stepped ahead and adopted image captioning to generate a description of the obstacle from the image. 
\subsubsection{Image captioning} 
Obstacle image classification has limitations. For example if there are multiple obstacle the classifier recognize only the apparent one. Classifier provides probability measurements of the multiple obstacles but difficult to incorporate in the application. Instead of classifying the obstacle image it is more design optimum to generate the caption of the image. Because the caption usually contains the description of the image, covering obstacles in the image. Image captioning performs nearly well to describe a general purpose image. The sidewalk obstacle image require accurate description so that the visually impaired is well aware of it. 

For the image captioning purpose, we investigated off-the-shelf APIs. Among those computer vision APIs, only Microsoft Cognitive service has the image captioning capability. Google, IBM BlueMix, and Clarifai have image tagging and concept generation service. Here we report the details of the experiment. 

In the first step, we initiated API calls from desktop computer. The calls were made using a single image to Google Vision, IBM BlueMix, Microsoft Cognitive service, and Clarifai. On an average Clarifai took less time and Cognitive service took a longer time to tag the image. Table \ref{tab:taggingtime} shows the result. The variation of the time is happening due to the free version of the APIs (we used the free version), the distance of the geo-location, or the data communication network delay.

\begin{table}
	\centering
	\caption{Captioning time for cloud API.}
	\label{tab:taggingtime}
	\begin{tabular}{|c|c|} 
		\hline
		API Host     & Time (ms)  \\
		\hline
		Clarifai & 381  \\
		\hline
		Google  & 617   \\
		\hline
		IBM BlueMix  & 838  \\
		\hline
		Microsoft & 1380    \\
		\hline
		
	\end{tabular}
\end{table}

Next step, we deployed the application in RPi3. Five sighted volunteers walked on the sidewalk with the device where there is wifi network available. They collected pictures of the sidewalk obstacle using the device in real-time. The image collection was performed to capture different variability that include, different diurnal circle, lighting condition shadow, low and bright sunlight, forward-backward motion of obstacle, moving obstacle, rotation, occlusion.

\begin{figure*}[h]
	\begin{center}
		\begin{tabular}{m{2cm} m{2cm} m{2cm} m{2cm} m{2cm}}
			
			\includegraphics[width=\linewidth,height=3cm]{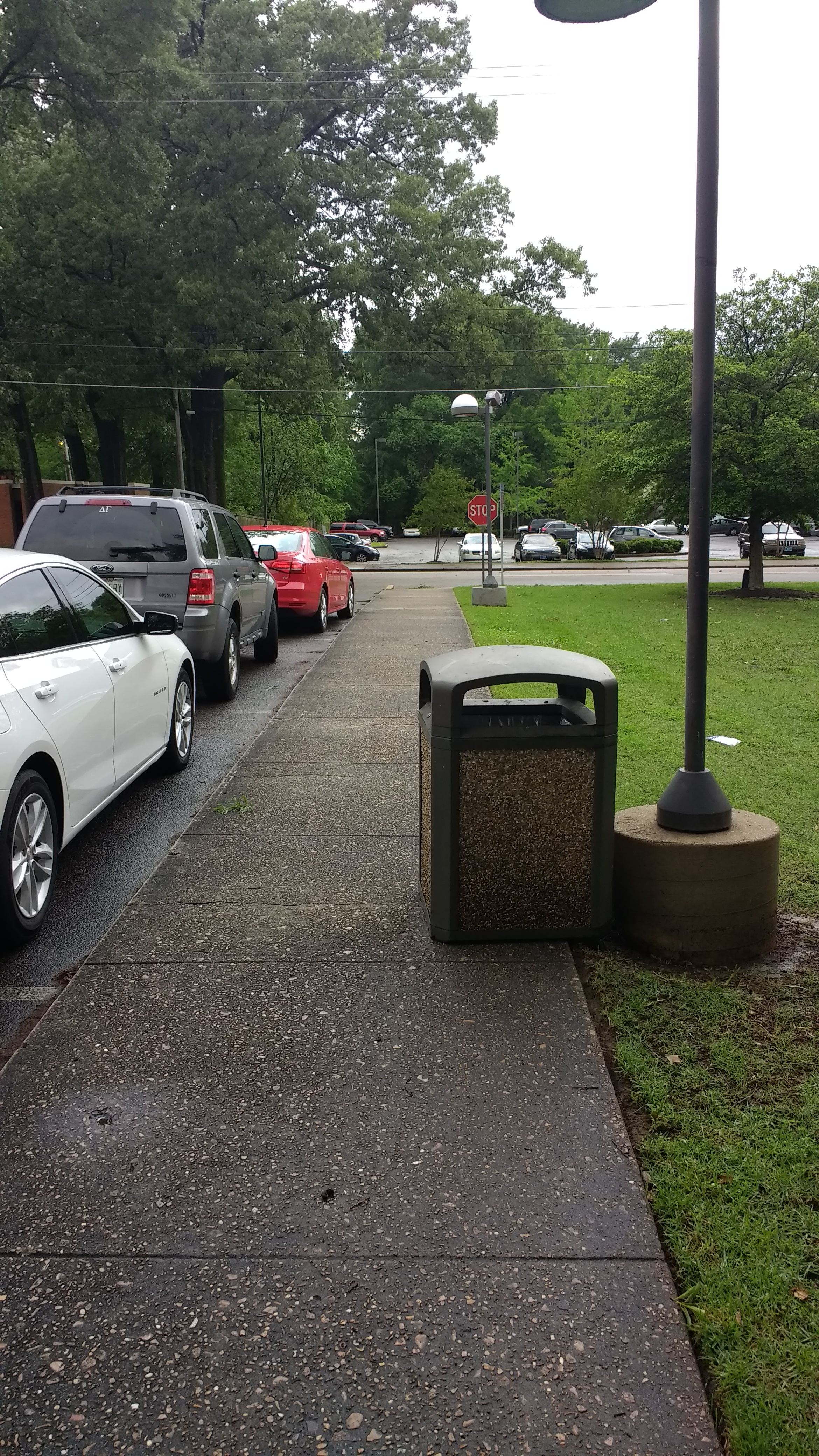}  &
			\includegraphics[width=\linewidth,height=3cm]{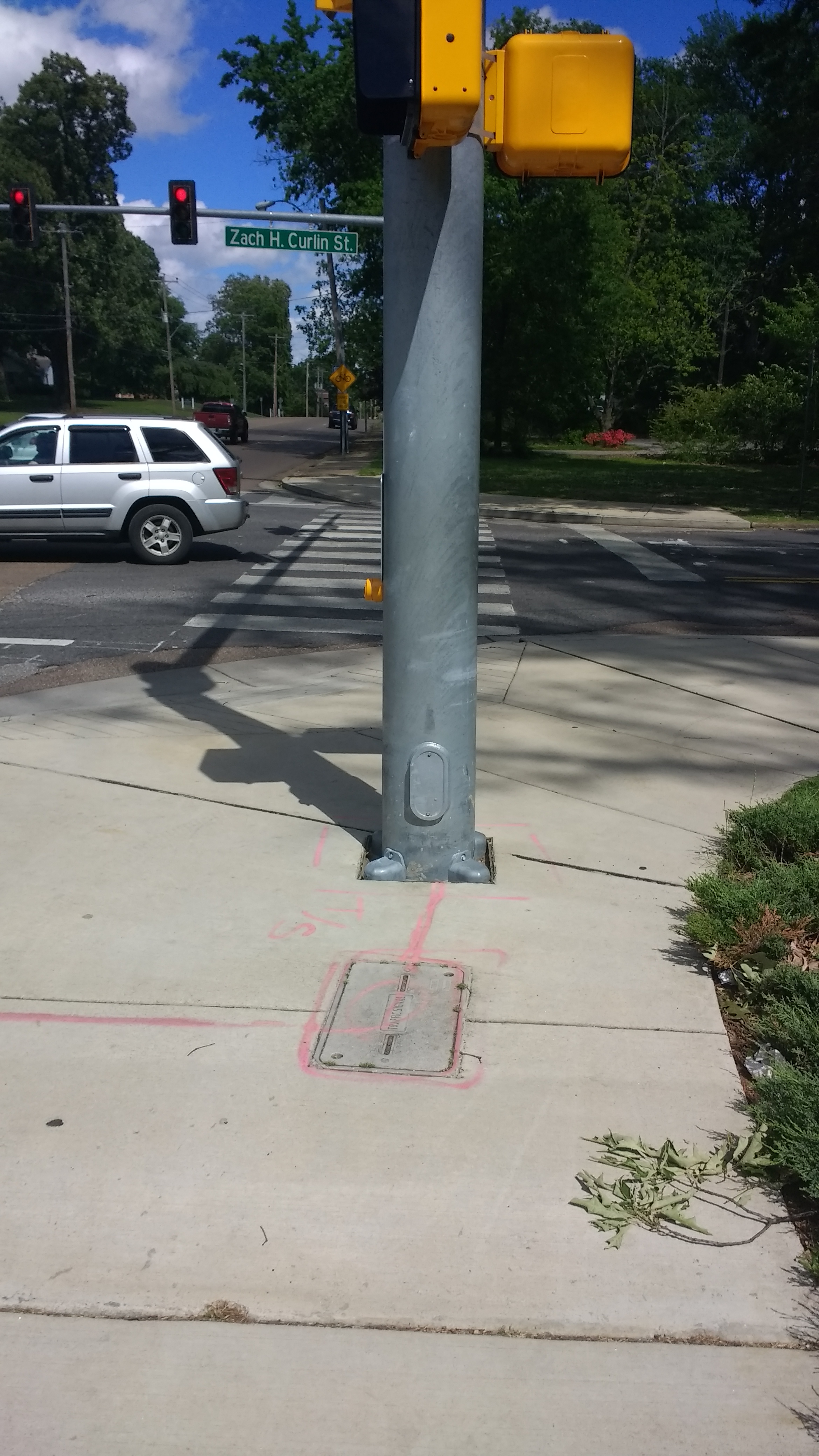}  &
			\includegraphics[width=\linewidth,height=3cm]{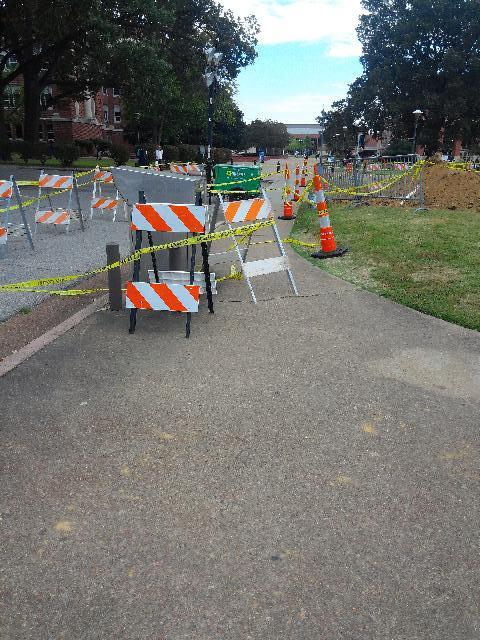}  &
			\includegraphics[width=\linewidth,height=3cm]{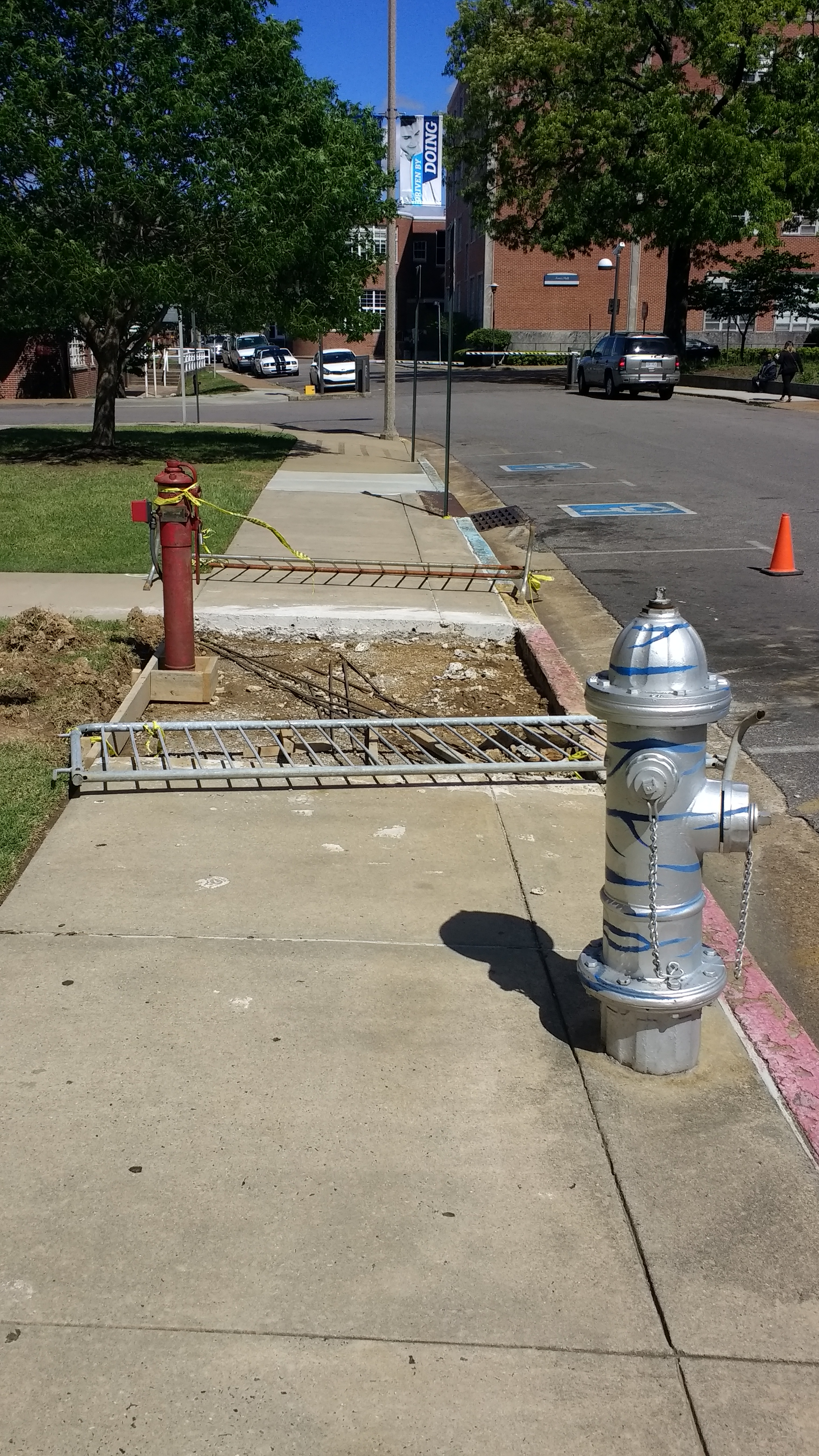}  &
			\includegraphics[width=\linewidth,height=3cm]{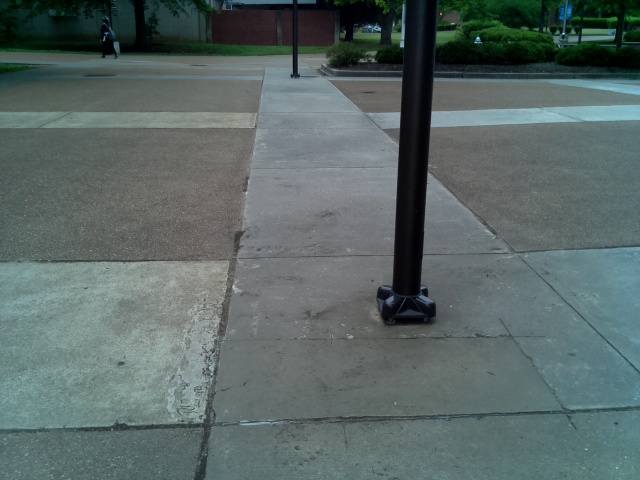}  \\
			
			(i) a car parked on the side of a road & 
			(ii) a traffic light sitting on the side of a road &
			(iii) a construction site &
			(iv) a fire hydrant on the side of the street & 
			(v) a pole sitting in the middle of a sidewalk \\
		\end{tabular}
		
	\end{center}
	\caption
	{Example of correct captions.}
	\label{fig:correctcaption}
\end{figure*}

It is observed that the captioning performs better when the image is frontalized, and the obstacle is occupied most space in the image. The performance reduces where there are multiple objects and obstacles. Some of the correct captioning (based on human judgment) cases are shown in Figure \ref{fig:correctcaption}. Most of the caption contains the word ``sitting'' and start with ``a''. This is because the model used to generate the caption was trained on a dataset which is basically the paired images with captions. 

\begin{figure*}[h]
	\begin{center}
		\begin{tabular}{m{2cm} m{2cm} m{2cm} m{2cm} m{2cm}}
			
			\includegraphics[width=\linewidth,height=3cm]{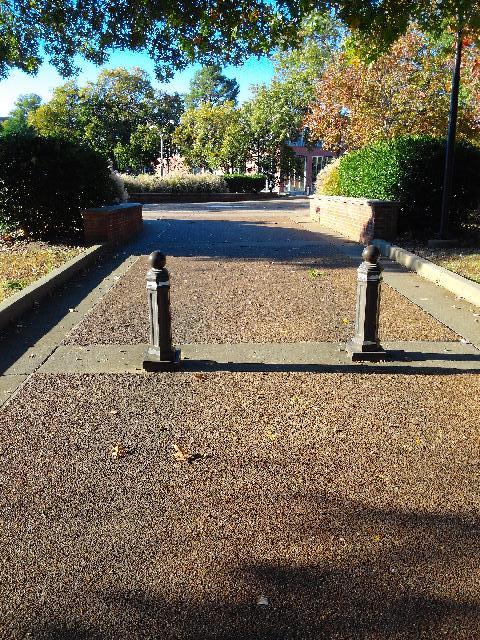}  &
			\includegraphics[width=\linewidth,height=3cm]{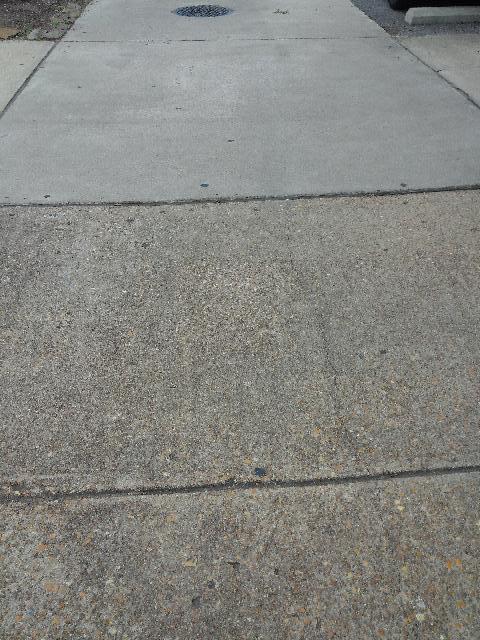}  &
			\includegraphics[width=\linewidth,height=3cm]{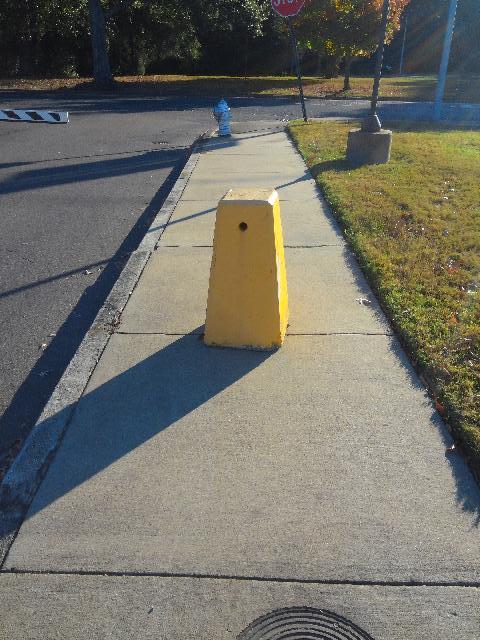}  &
			\includegraphics[width=\linewidth,height=3cm]{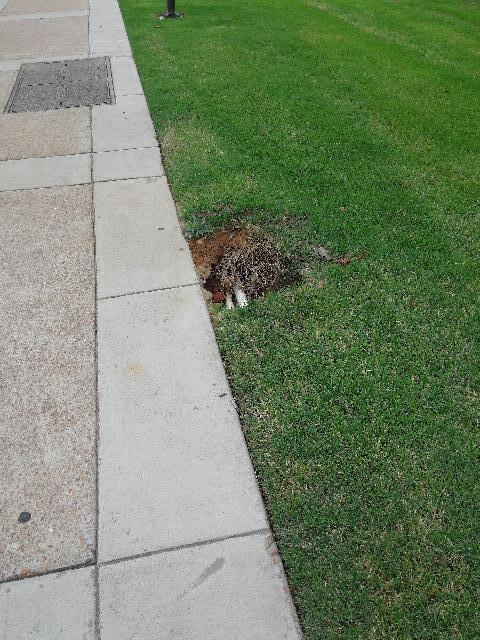}  &
			\includegraphics[width=\linewidth,height=3cm]{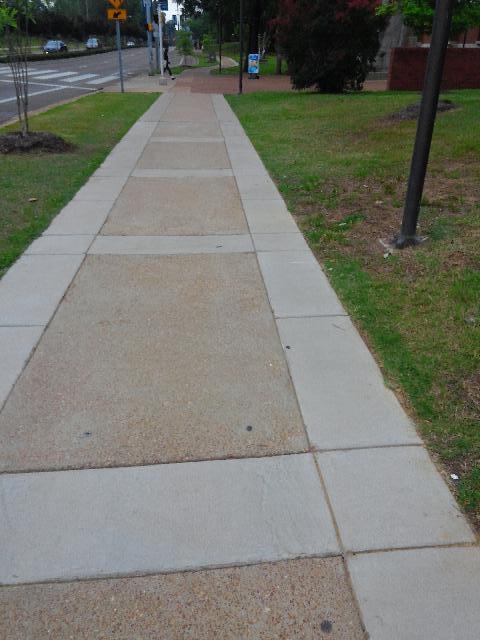}  \\
			
			(i) a bench on a sidewalk & 
			(ii) a fire hydrant on the sidewalk &
			(iii) the side of a road &
			(iv) a dog laying on a sidewalk & 
			(v) a fire hydrant on the sidewalk\\
		\end{tabular}
		
	\end{center}
	\caption
	{Example of incorrect captions.}
	\label{fig:failedcase}
\end{figure*}

There are a few situations where the visual description was not correct. Examples are in Figure \ref{fig:failedcase}. In the picture (i), the bollards are thought to be bench, and there is no obstacle in the picture (ii), but caption talks about fire hydrant. Moreover, in the picture (iv), the pothole looks like a dog laying and in picture (v),  the pole is described as fire-hydrant. The image (v) is not bad because even the pole is described as fire-hydrant. At least from this description the blind person may get alerted.

\begin{figure*}[h]
	\begin{center}
		\begin{tabular}{m{2cm} m{2cm} m{2cm} m{2cm} m{2.5cm}}
			
			\includegraphics[width=\linewidth,height=3cm]{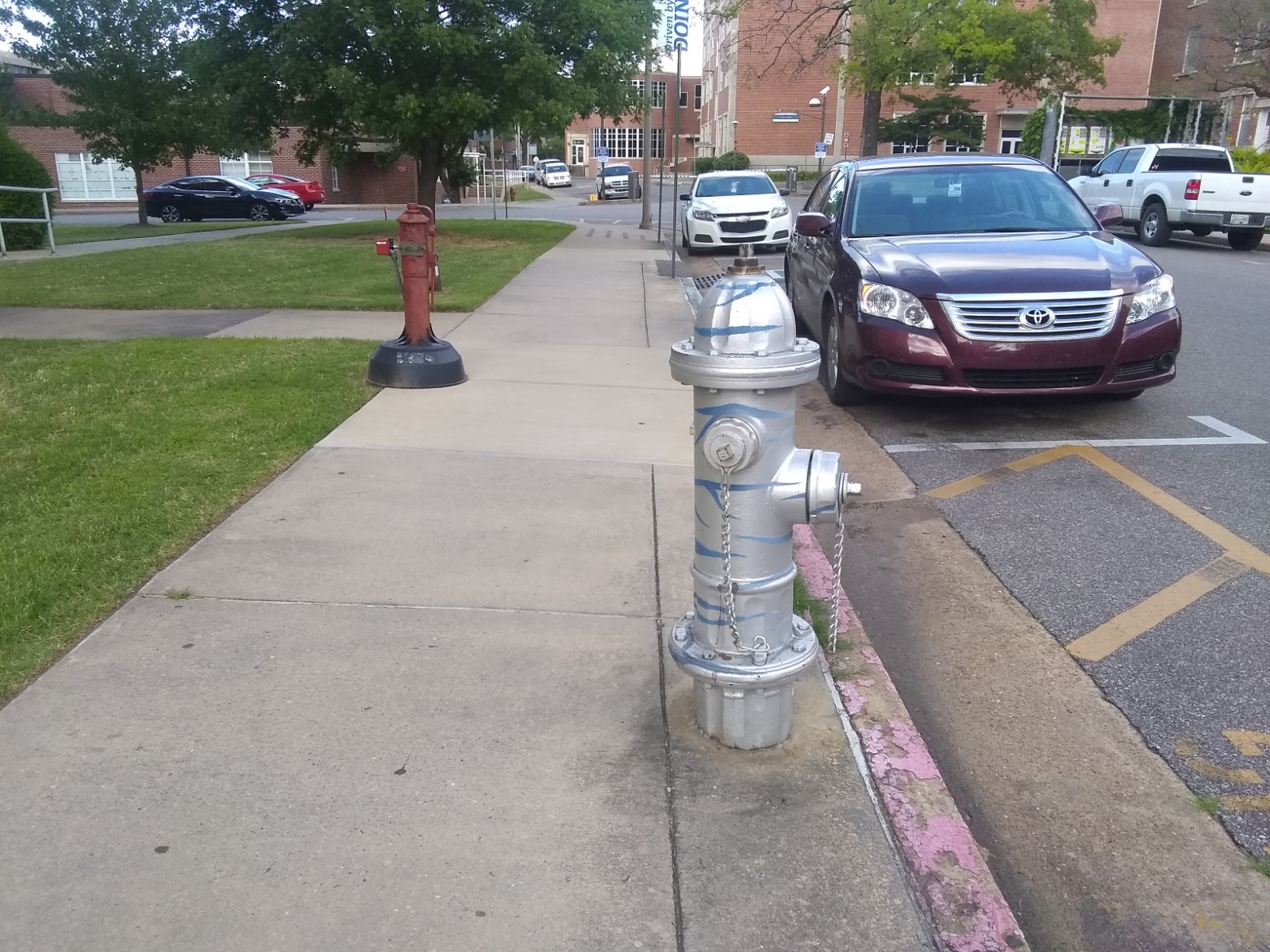}  &
			\includegraphics[width=\linewidth,height=3cm]{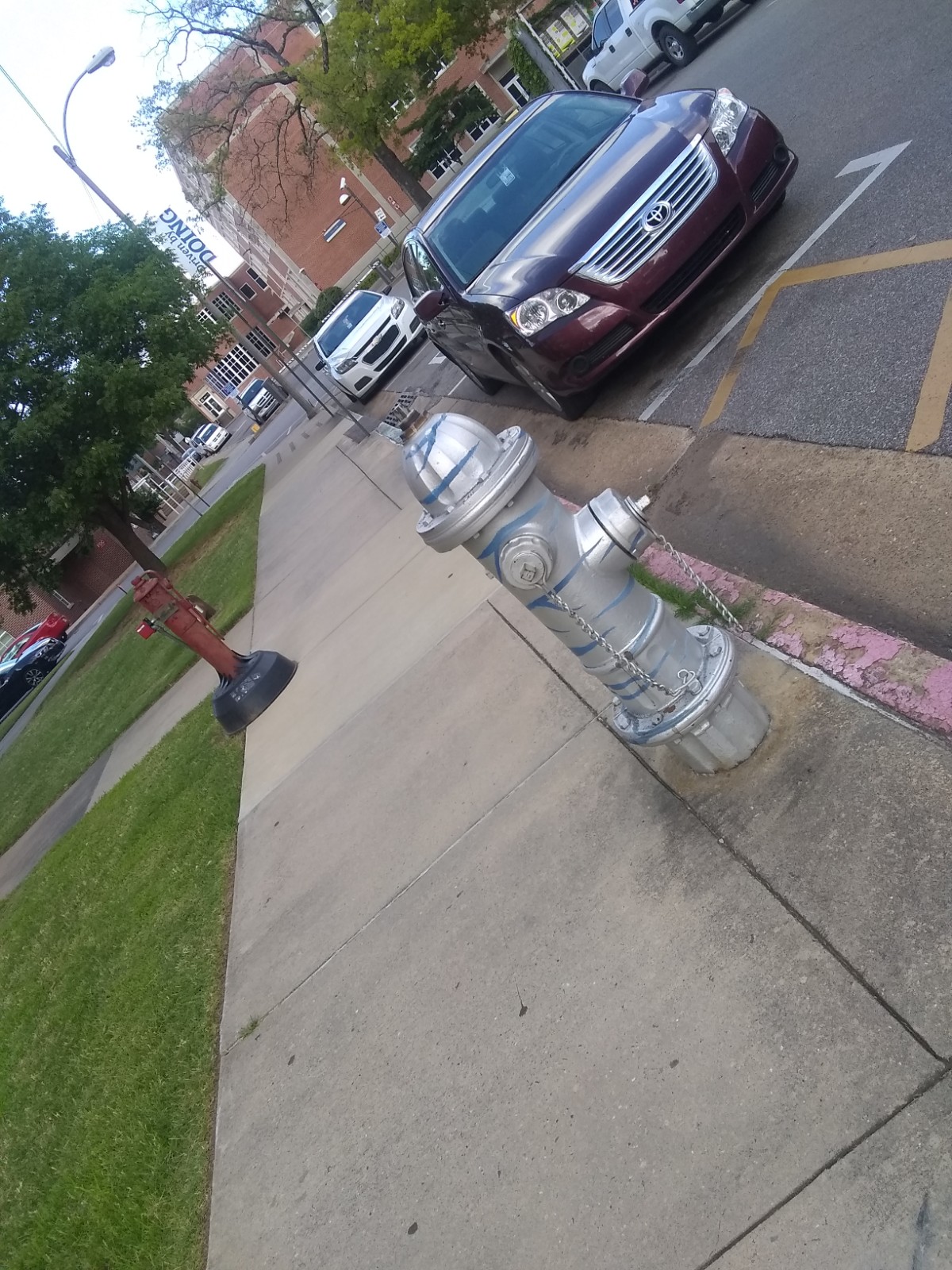}  &
			\includegraphics[width=\linewidth,height=3cm]{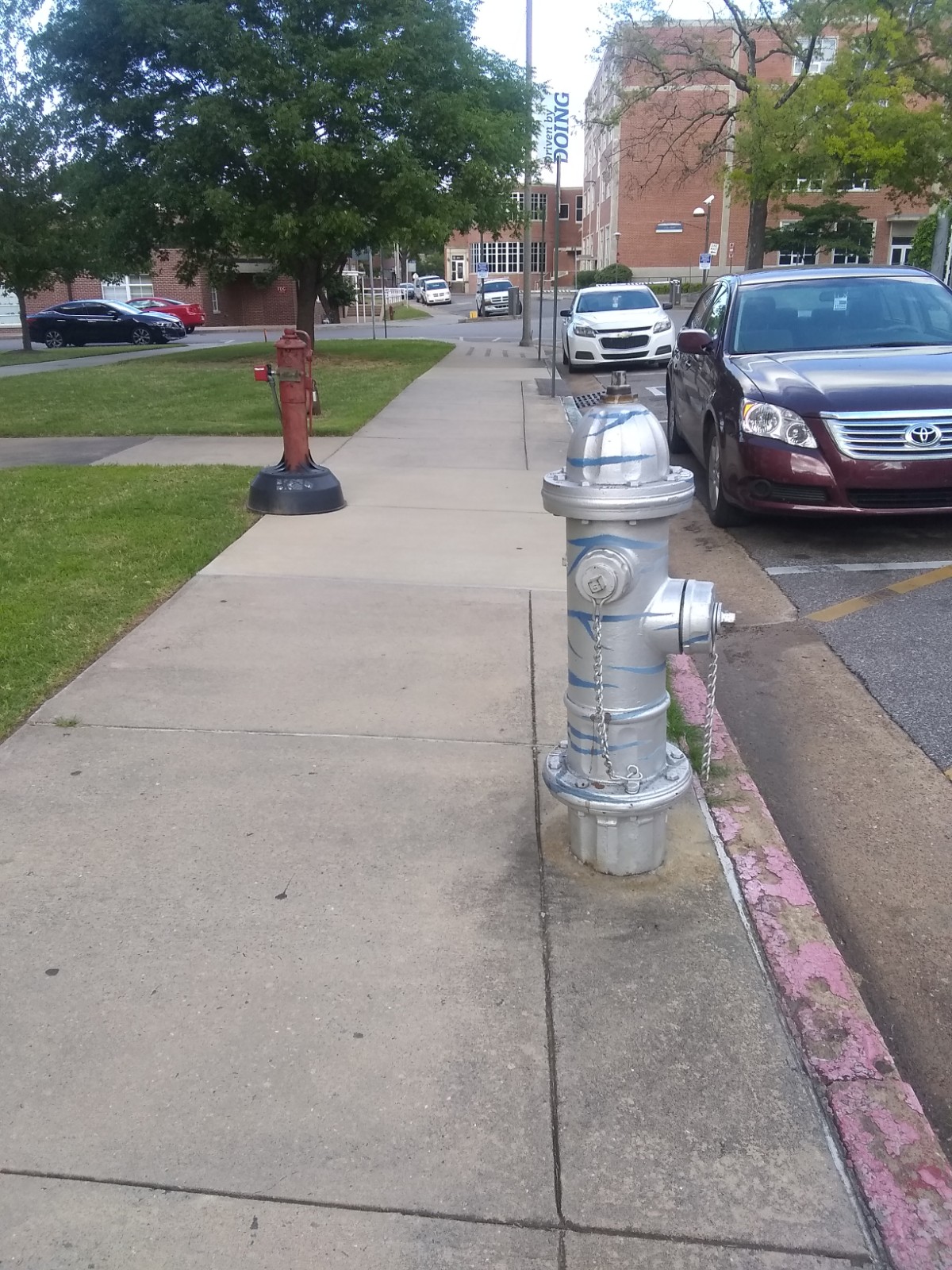}  &
			\includegraphics[width=\linewidth,height=3cm]{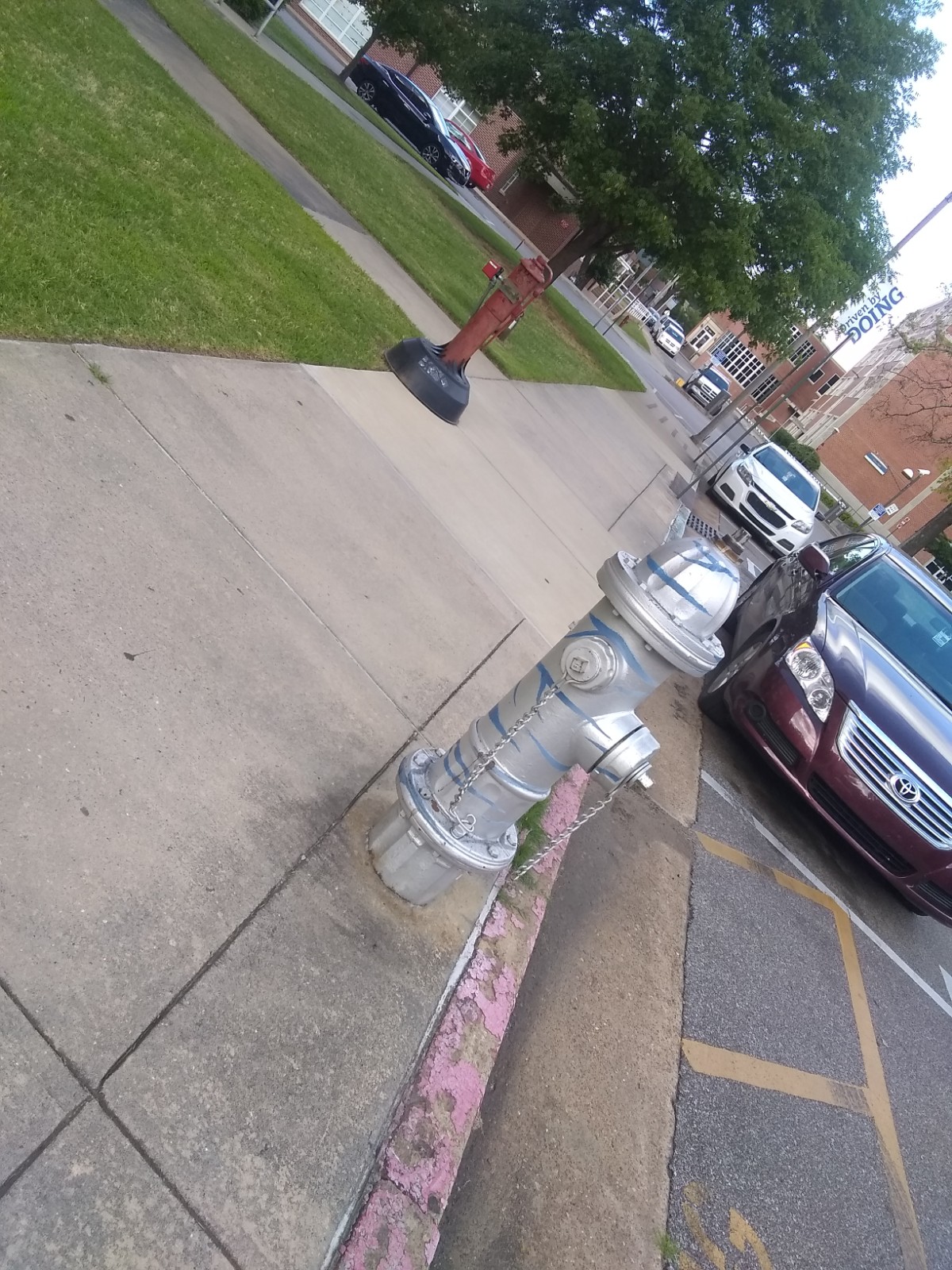}  &
			\includegraphics[width=\linewidth,height=3cm]{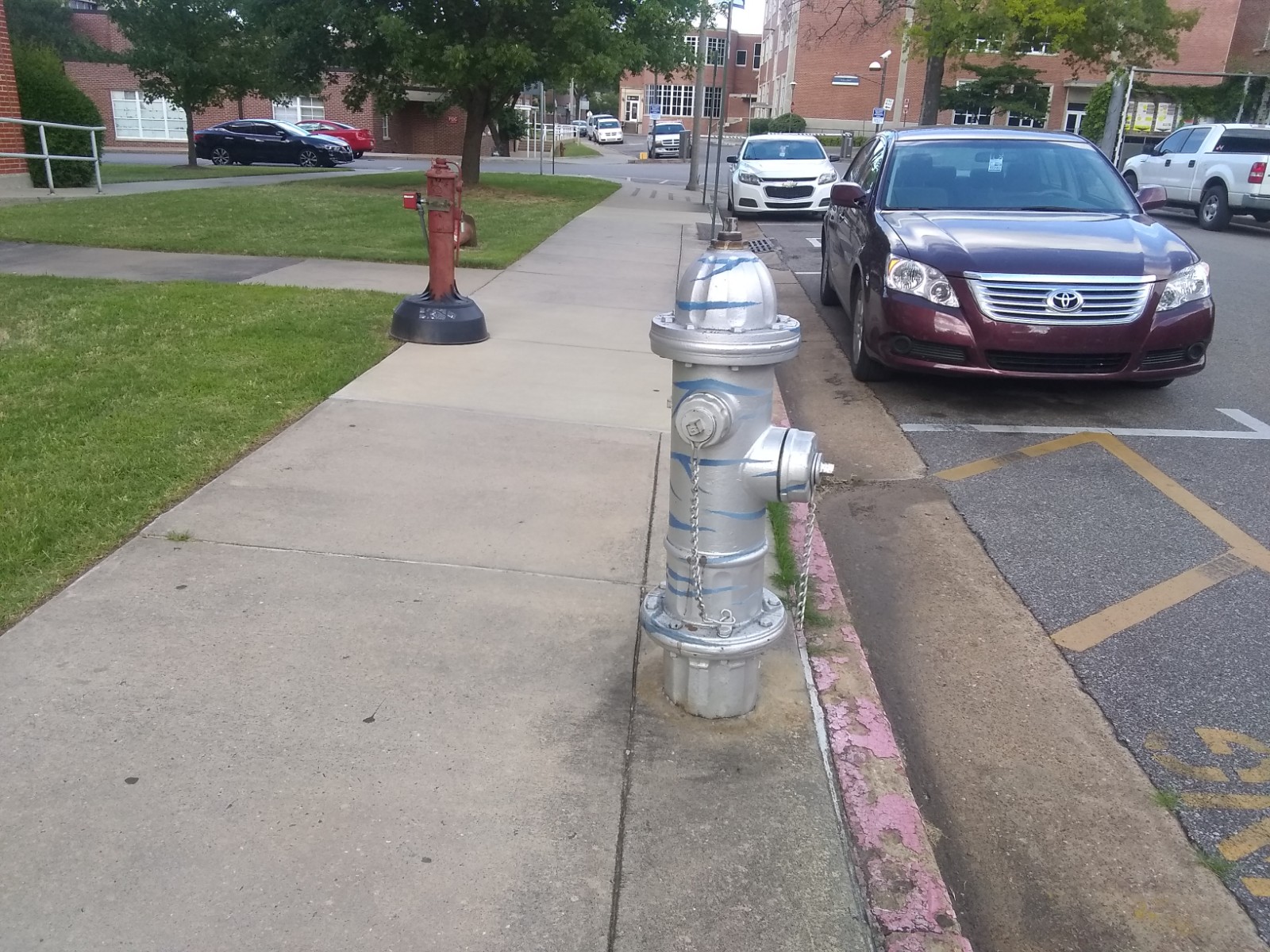}  \\
			
			(i) person walking down a sidewalk & 
			(ii) a sandwich on a sidewalk &
			(iii) a fire hydrant on the side of a fence &
			(iv) a man laying on a sidewalk & 
			(v) a cat is standing on a sidewalk\\
		\end{tabular}
		
	\end{center}
	\caption
	{Incorrect captions due to rotation.}
	\label{fig:rotated}
\end{figure*}

\textit{Rotation.} To observe the rotational case, we manually rotated the camera to get $90$ degrees rotation and $45$ degrees rotation we obtained through image rotation function. Figure \ref{fig:rotated} shows some captions of rotated image. The captions generated for the rotated images does not describe the image properly. The $45$ degrees rotated image captions is not correct due to the dark portion due to the rotation. It is promising that at least the model was able to identify the sidewalk in all four rotated cases, though the captions are not relevant.   

\begin{figure*}
	\begin{center}
		\begin{tabular}{m{2.5cm} m{2.5cm} m{2.5cm} m{2.5cm}}
			
			\includegraphics[width=\linewidth,height=3cm]{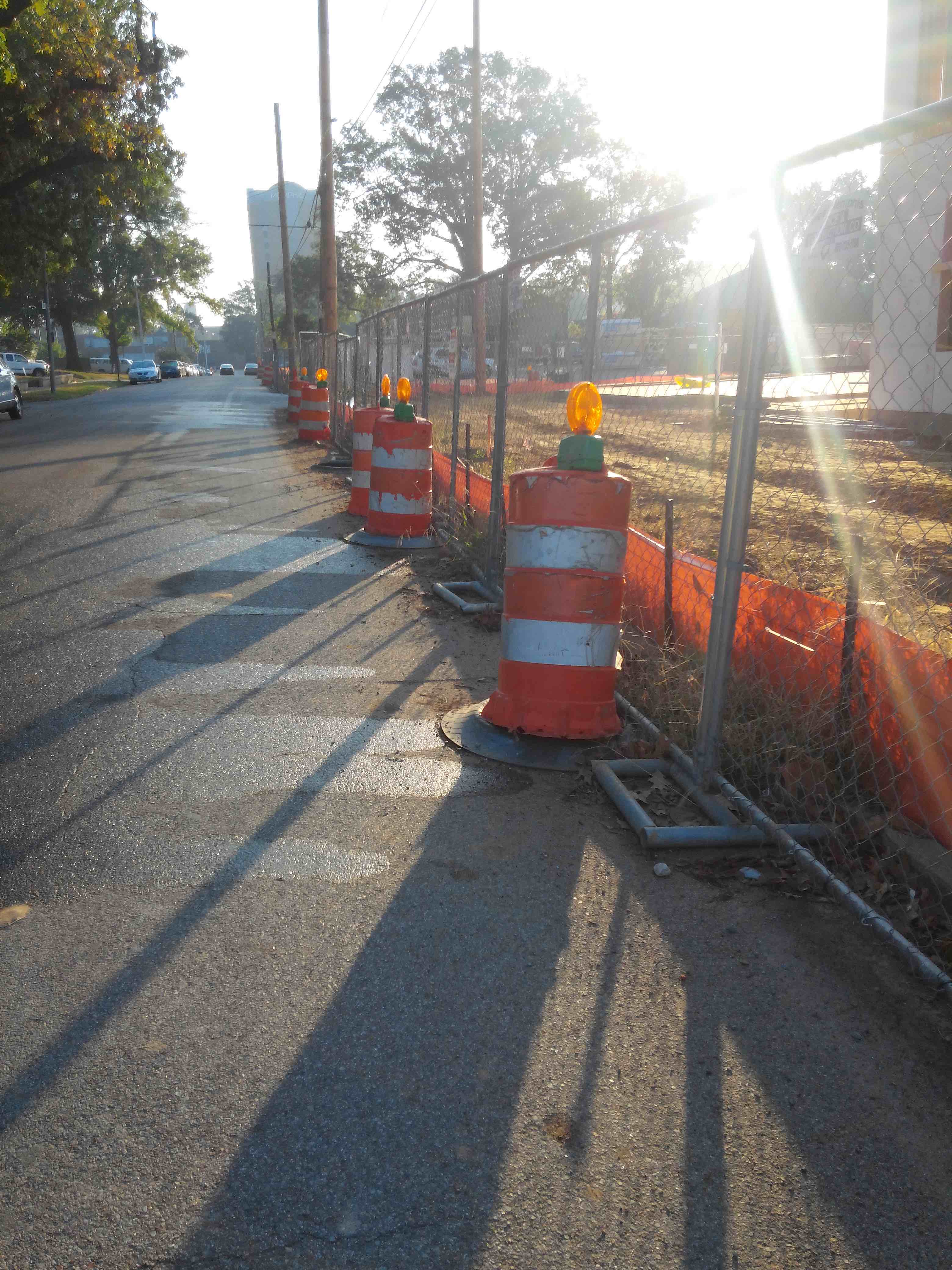}  &
			\includegraphics[width=\linewidth,height=3cm]{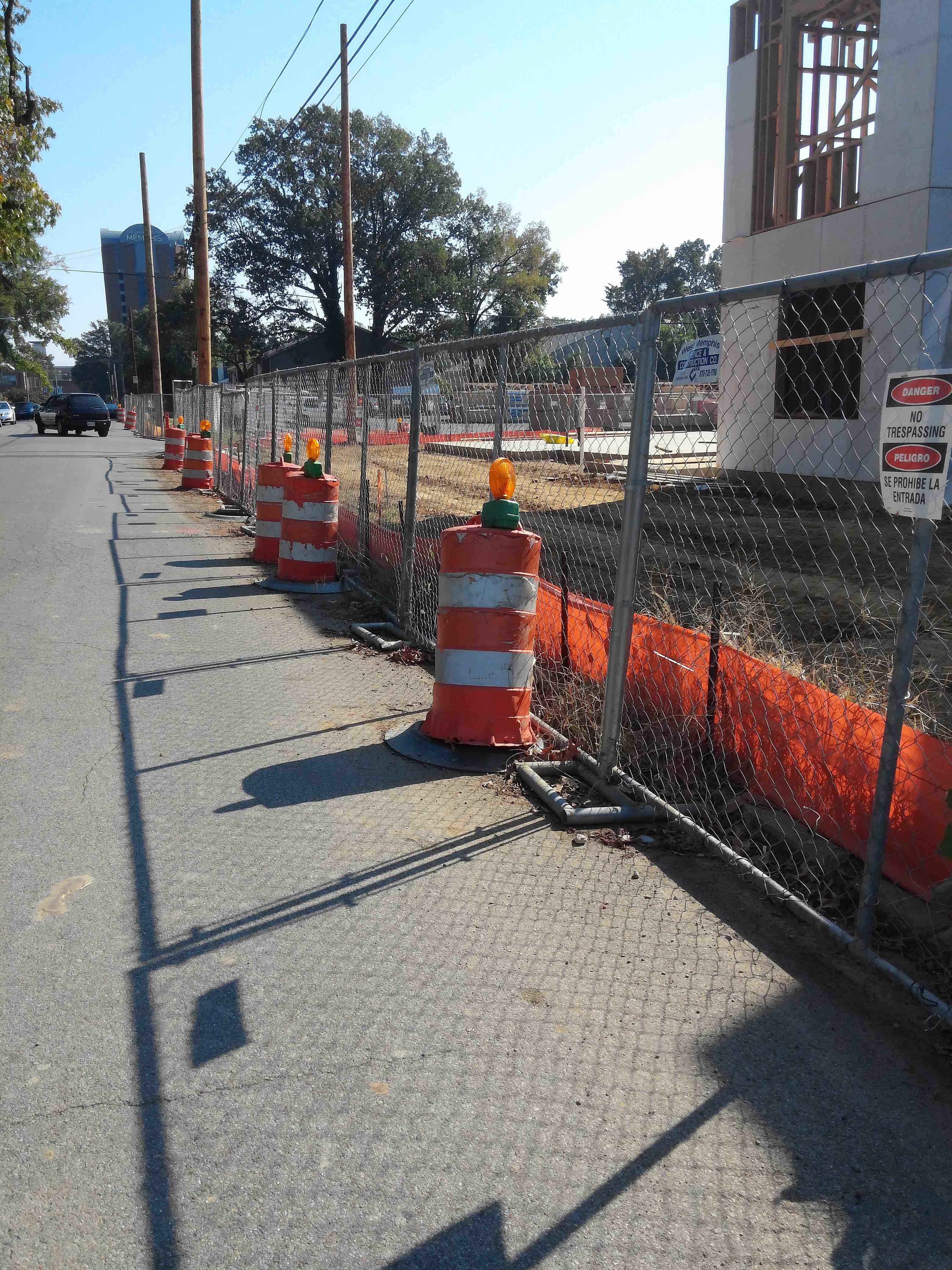} &
			\includegraphics[width=\linewidth,height=3cm]{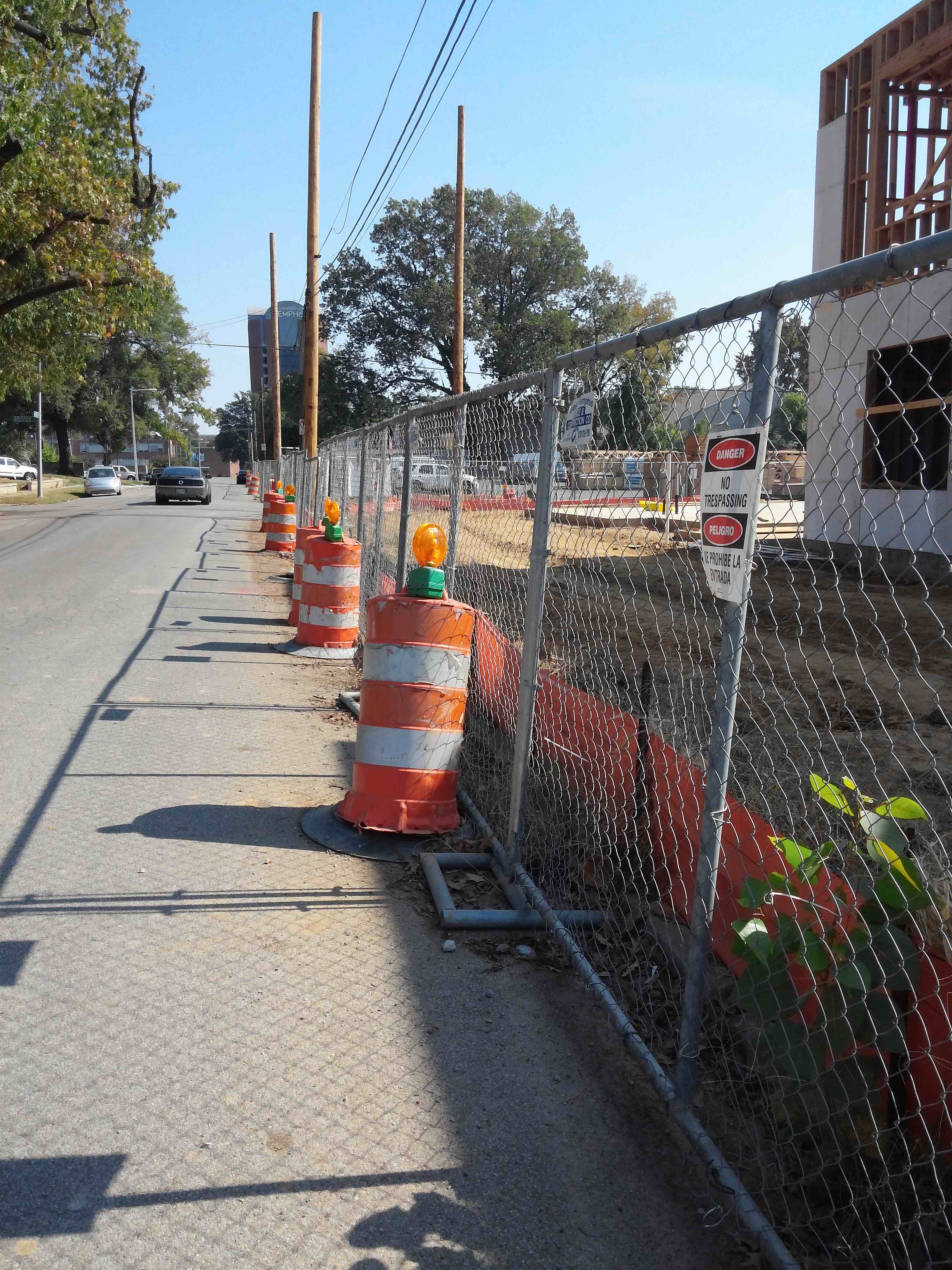}  &
			\includegraphics[width=\linewidth,height=3cm]{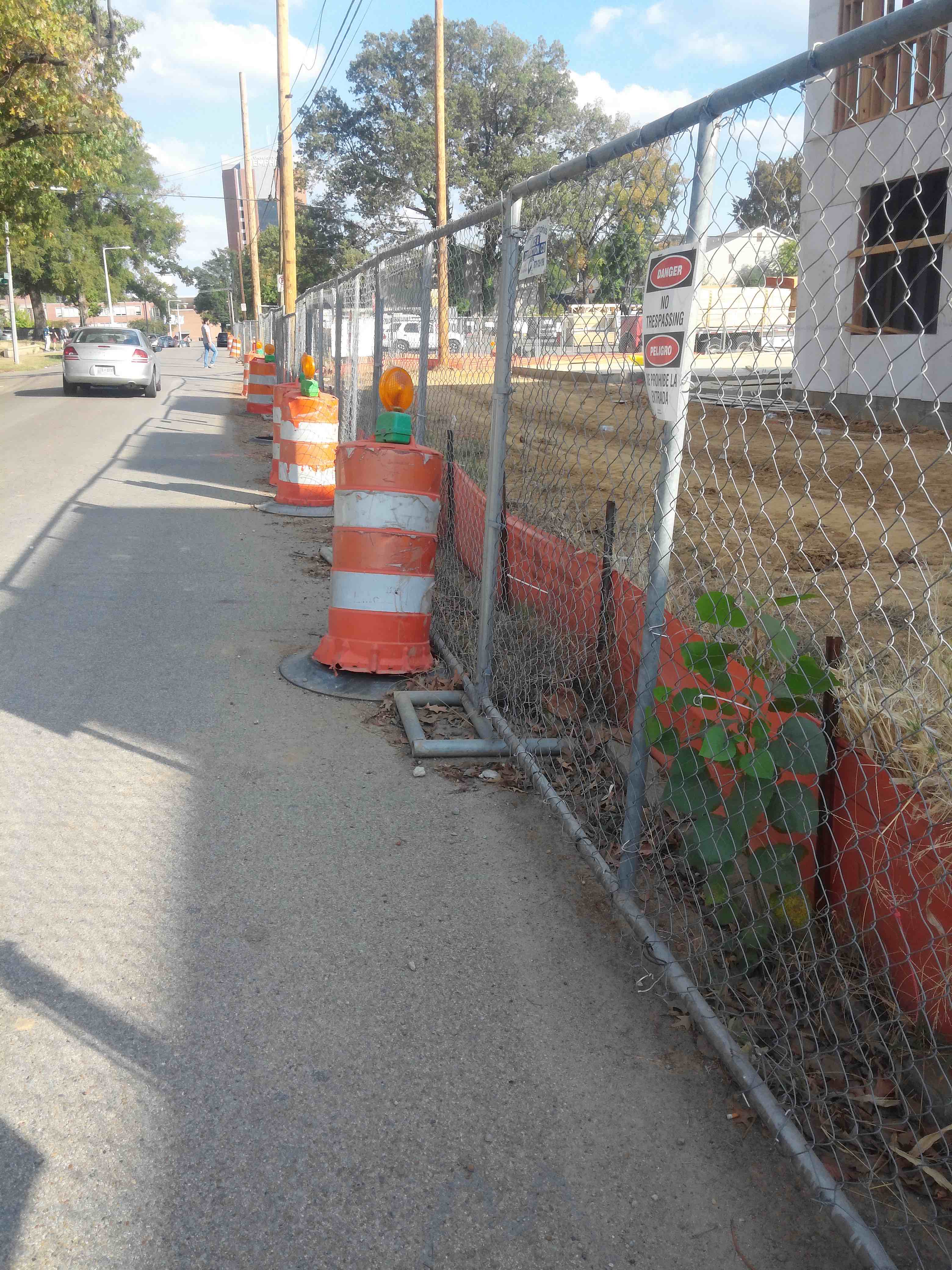}  \\
			
			(i) a red train traveling down tracks next to a fire hydrant & 
			(ii) a construction site &
			(iii) a fire hydrant on the side of a fence &
			(iv) a construction sign on the side of a road \\
			
		\end{tabular}
		
	\end{center}
	\caption
	{Example of diurnal effect (pictures taken at 8AM, 11AM, 1PM, and 4PM of the same obstacle).}
	\label{fig:diurnaleffect}
\end{figure*}

\textit{Diurnal effect.} The captioning task for image also was not correct where there is shadow due to diurnal effect. The images in Figure \ref{fig:diurnaleffect} are taken from a certain place of same obstacle at different time of the day. However, the captions are different and sometimes very irrelevant. The first image talks about a ``red train'', but there is no train in the image. Probably the long shadow looked like a train. Other three images are somewhat relevant to the construction site which is apparent.

Describing an image from a caption to visually impaired was more informative than image classification we learnt from users' feedback. It was because users obtained a description of a scene. We observed that the description is far less than accurate for dynamic obstacles (e.g., the obstacles which are moving, cars, bicycles). 
At this stage we revised the problem and come up with an innovative idea that is to find ``free path'' instead of ``modeling diverse obstacles''. Implementing this idea falls under ``system thinking'' which is the next subsection. Moreover, it is noticed that the image-based solutions suffer from high computation cost and heavily depend on visible light \cite{ahmed2017optimization,ahmed2018image,ahmed2018interactive} which is another motivation for the revision of the problem.

\subsection{System thinking}
The idea of finding and incorporating ``free path'' in assistive system to avoid obstacle has three components
\begin{itemize}
    \item Modeling free path in a simulated robotic environment
    \item Train an agent to avoid obstacles using free path, we name this Sidewalk Obstacle Avoidance Agent (SOAA)
    \item Train another agent responsible for natural conversation, which is Sidewalk Obstacle Conversation Agent (SOCA).
\end{itemize}
Combining these system components into a single implementation gives augmented guiding (AG).


\subsubsection{Free-path definition}
Generally the sidewalk consists of transitive and intransitive obstacles. The transitive obstacles have velocity. The visually impaired person walking on the sidewalk has a certain speed. The mean comfortable walking speed of adult (aged between 20 to 70 years) ranges approximately from $100$ cm/s to $150$ cm/s.  \cite{walkingspeed}.  

Suppose $ \chi = (M,d)$ is a discrete metric space from Euclidean space $\mathbb{R}^n$, where $M \subset \mathbb{R}^n$ is the set of points and $d$ is the distance metric. The density of $M$ in the ambient Euclidean space may not be uniform due to perspective distortion. There exist a set of functions $f$ that take $\chi$ as input and produce clusters satisfying a set of constraints (e.g., points at a given neighborhood distance or color) \cite{qi_pointnet:_2016}. 

In the given $\chi$ the \emph{Free Path} is defined as $f(\chi)=\phi$ which indicates there is no obstacle along the direction of interest. If $f(\chi)=C$ where $C$ is a set of clusters in $\chi$. The \emph{Threat Level} $t$ is inversely proportional to the distance of the cluster $c_i$ ( $C \in \{c_1,c_2,...,c_i\}$), that is $t \propto \frac{1}{d_i}$ \cite{morales_combined_2017,lee_method_2017}. 

As an alternative of the image-based solution finding ``free path'' is more easier from point cloud (PC). Because each points in PC contain distance information. PC is a set of data points in space \cite{lee2017method}. The PC is constructed from laser technology-based camera (e.g. Microsoft Kinect) and contains RGB colors with depth information (RGB-D). Additional advantage is that the PC is not extremely dependent on visible light. In addition preparing simulation for ``free path'' and adopting that model in real world is easily achievable. The simulation also provides better understanding of a problem and feasibility of proposed solution.  

\begin{figure}[htb]
	\begin{center}
		\includegraphics[width=0.5\linewidth]{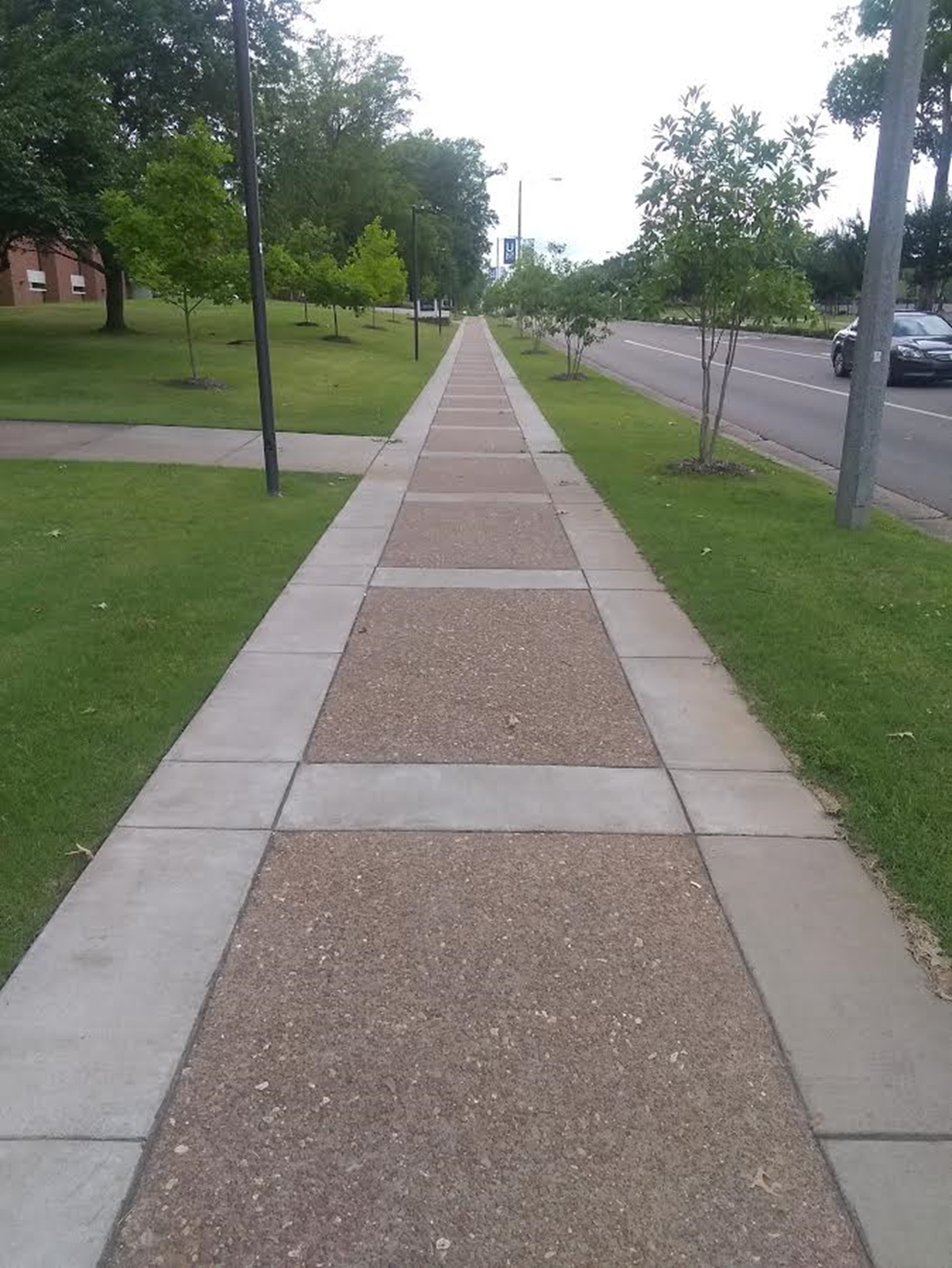}
	\end{center}
	\caption{Typical sidewalk.}
	\label{fig:typicalsidewalk}
\end{figure}

A typical sidewalk is presented in Fig. \ref{fig:typicalsidewalk}. The sidewalk itself is an environment with many different states with obstacle where a robot takes action and either gets rewards or punished. We have defined the environment, state, action, and reward in terms of sidewalk in the following manner. 

\textit{Environment:} The sidewalk environment consists of static and dynamic obstacles. The static obstacle does not move whereas the dynamic obstacle moves. The sidewalk has a curb and it has brick pavement. There is grass beside the sidewalk which is different in color than the sidewalk itself. 

\textit{State:} The state space is a set of all possible relative positions of agents and the obstacles on the sidewalk. That is why the number of states is infinite. The agent finds useful information from the states to make the right action.    

\textit{Action:} There are five actions namely stop, left, forward, right, and backward. The agent encounters infinite number of states and takes one of these actions in the action space set. 

\textit{Reward:} If an action performed by the agent causes collision, then the reward is $-1$. The agent keeps getting $+1$ as a reward until there is no collision.

The sidewalk with an environment, states, actions, and rewards resembles a markov decision problem (MDP). Thus this can be solved using reinforcement learning (RL).

\subsubsection{Sidewalk Obstacle Avoidance Agent (SOAA)}
The SOAA is the combination of camera and lidar sensors. For the convenient the SOAA is plugged in to a robot and simulated in Gazebo environment. Gazebo environment is embedded with physics engine and easy to place obstacles. Moreover it is easy to implement learning algorithm in Gazebo. 

\begin{figure}
	\begin{center}
		\begin{tabular}{m{4cm} m{4cm}}
		  \includegraphics[width=\linewidth]{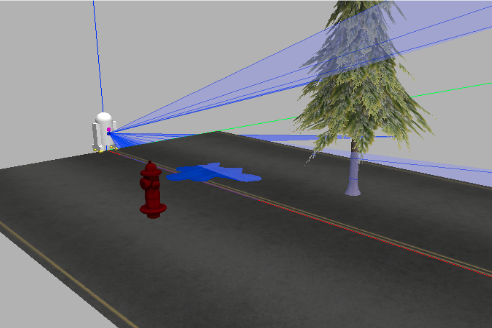} &
		  \includegraphics[width=\linewidth]{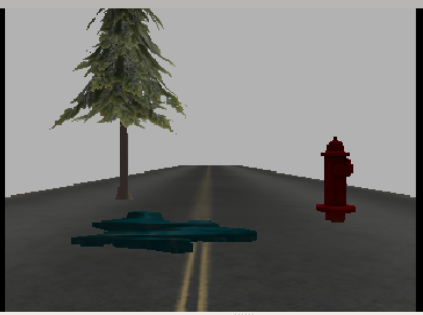} \\
		   (a) Gazebo simulation & (b) What robot sees \\
		  \includegraphics[width=\linewidth]{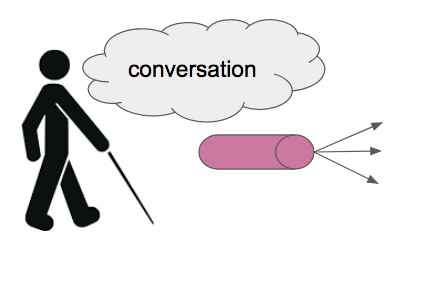} &
		  \includegraphics[width=\linewidth]{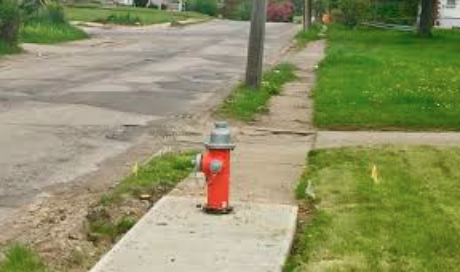} \\
		 
		  (c) Real sidewalk & (d) What SOAA sees 
	    \end{tabular}
	\end{center}
	\caption{Analogy with SOAA and Gazebo simulation.}
	\label{fig:analogyagtgazebo}
\end{figure}

In the gazebo environment we implemented the reinforcement learning where there are obstacles and robot tries to avoid finding the free path. The robot feels the obstacles through lidar and sees it through the camera. It also learns to decide which direction to turn to avoid the obstacle Fig \ref{fig:analogyagtgazebo}.

\begin{figure}[htbp]
	\begin{center}
			\includegraphics[width=0.8\linewidth]{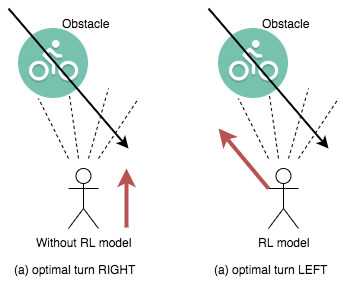} 
	\end{center}
	\caption{AG learns optimal turn through reinforcement learning.}
	\label{fig:optimalturn}
\end{figure}

There is a question, why reinforcement learning necessary. We choose reinforcement learning for the dynamic nature of the obstacles. The static obstacles robot can avoid with rule based model. But to avoid the dynamic (e.g., moving) obstacles the robot has to learn from the experience and decide optimal turn Fig \ref{fig:optimalturn}. This is the reason for choosing reinforcement learning.     

Once we train the reinforcement model, this is integrated with the SOAA. The SOAA does not decide the turn or move, it gives the ambient information about the obstacle and the person decides which direction to move. Thus the decision process for robot is replaced by the decision process of human being. The aspects of the problem to be reinforcement learning are 

\begin{itemize}
	\item Different actions yield different rewards. For example, when trying to avoid obstacle in a sidewalk, going left may lead to an avoidance, whereas going right may occur collision.
	\item Reward for an action is conditional on the state of the environment. From the above Fig \ref{fig:optimalturn}, going left may be ideal at a certain position in the path, but not at others.
	\item Rewards are delayed over time. This just means that even if going left (Fig \ref{fig:optimalturn}) is the right thing to do, we may not know it till the obstacle is complete out of sight.	
\end{itemize}

\textbf{The reinforcement model} is built through the policy gradient learning. The lidar senses the obstacle and the SOAA position relative to the static and dynamic obstacle is the \textbf{State}. \textbf{Actions} are every desired movement of the SOAA in the simulated environment. Finally the \textbf{Reward} is the obstacle avoidance. Ideally the Q-learning algorithm should give positive reward when the SOAA reaches at the target. The target in this case is the avoidance obstacle. 

The deployed model in SOAA monitors the states of the SOAA and talks about it. The person takes action. This is the production environment. The SOAA takes picture of the obstacle and annotates the picture. The top 5 components are mentioned to the user along with the relative state.     

\subsubsection{Sidewalk Obstacle Conversation Agent (SOCA)}
In daily life, any matter not apparent to the user becomes more transparent through the conversation. That is why the teachers request students to ask questions, and the managers ask the employee to ask questions. Through the conversation, the real scenario becomes evident. 

In this research, we are adopting this concept. The user communicates with the agent, and the agent talks about what it sees ahead. Through the conversation, the ambiance become more apparent to the user. The agent mentions any obstacle on the walkway to the user. How to avoid that obstacle depends on the user. The SOAA device will not command to do a particular action. Instead, the user decides the next action based on the conversation. This conversational agent is an active interface. 

For the basic understanding of the conversational agent we define few keywords.

\textit{Intent:} The intent is the end meaning of what the user is trying to say. For example, if the user says, ``Find the fire hydrant'' the intent can be classified as to find obstacle.

\textit{Entity:} An entity is to extract useful information from the user input. From the example above, ``Find the fire hydrant'' the entities extracted should be the \textit{name} of the obstacle. The name, for example, is a fire hydrant. 

\textit{Stories:} Stories define the sample interaction between the user and the conversational agent connecting intent and action performed by the agent. In the example above agent got the intent of finding the obstacle and entities like the name of the obstacle, but still, there is an entity missing - how far should it look. That would make the next action from the agent.

\textit{Actions:} Actions are the operations performed by the agent. It could be either asking for some more details to get all the entities or integrating with some APIs or querying the RL model to get any information.  

\textit{Templates:} The templates are the sample replies from the agent which can be used as actions.

\begin{figure}
	\begin{center}
		\includegraphics[width=0.8\linewidth]{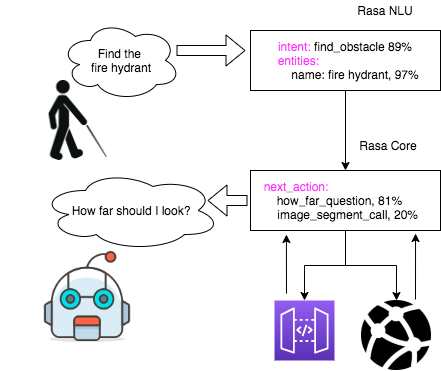} 
	\end{center}
	\caption{Block diagram of Rasa NLU and Rasa Core.}
	\label{fig:rasablockdiagram}
\end{figure}

The conversational agent, a software system, enables a user to talk with it in natural language. RASA, an open-source machine learning framework, serves as the engine of the conversational agent. It is easy to customize. We can build, deploy, or host RASA internally in our server or environment with complete control.  
Confidential conversation data cannot be shared with third party. The majority of the conversational agent tools available are cloud-based and provide software as a service. We cannot run them internally in our environment, and we need to send data to the third party. With RASA, there is no such issue.

The RASA comprises of two main components \textit{Rasa NLU} and \textit{Rasa Core}. Rasa NLU is a library for natural language understanding (NLU), which does the classification of intent and extract the entity from the user input and helps the agent to understand what the user is saying. Rasa Core, on the other hand, is a conversational agent framework with machine learning-based dialogue management capabilities. It takes the structured input from the NLU and predicts the next possible best action using a probabilistic model like long short-term memory (LSTM) recurrent neural network. Rasa NLU and Rasa Core are independent, and we can use NLU without Core, and vice versa. But using both NLU and Core enhance performance. A block diagram of RASA is shown in figure \ref{fig:rasablockdiagram}. 

Three types of files are necessary to train Rasa NLU. NLU training file, Stories file, and Domain file. The training file contains some training data with user inputs along with the mapping of intents and entities present in each of them. The more varying examples we provide, better the agent's NLU capabilities become. Stories file contains sample future interactions between the user and the agent. Rasa Core creates a probable model of interaction from each story.
The Domain file lists all the intents, entities, actions, templates, and some more information. The conversational data obtained from the WoZ experiment is converted to text and processed to create the above-mentioned training files. The training files are stored in markdown format. Samples form an NLU file is presented in listing \ref{listing:nlu}.    

\begin{verbatim}

## intent:greet
- hey
- hello
- are you there?
- are you ready?
- ready?

## intent:greet_ask
- Yes ready, are you ready?
- Ready, want to start?.
- I am here, start walking?

## intent:greet_normal
- yes
- yap
- let's go

## intent:find_obstacle
- Find [obstacle](obstacle)?
- What is [there](obstacle)?
- What is [that] (obstacle)?
- Do you see [anything](obstacle)?
- [There] (obstacle)?
- [Here] (obstacle)?
- This [way](obstacle)?
- That [way](obstacle)?

## intent:find_distance
- [Where](distance)?
- How [far](distance)?
- How long to [reach](distance)?
- Is it [close](distance)?
- Is it very [close](distance)?

## intent:bye
- bye, let me know
- bye now
- i am here, bye


\end{verbatim}

\begin{figure}
	\begin{center}
		\includegraphics[width=0.6\linewidth]{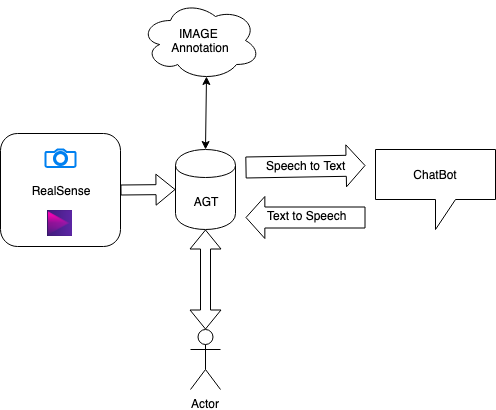} 
	\end{center}
	\caption{Block diagram of SOAA.}
	\label{fig:agtblockdiagram}
\end{figure}


\subsection{Assistive thinking}
SOCA is part of assistive thinking. We never want to identify a visually impaired as visually impaired, that is why the combination of SOAA and SOCA must be 
\begin{itemize}
    \item Wearable 
    \item Light weight
    \item Private conversation 
\end{itemize}

Add details of wearable, light weight, privacy.

\section{Evaluation}
\subsection{Evaluation of the models}
\begin{figure}[h!]
	\begin{center}
		\begin{tabular}{ccc}
			
			\includegraphics[width=0.3\linewidth]{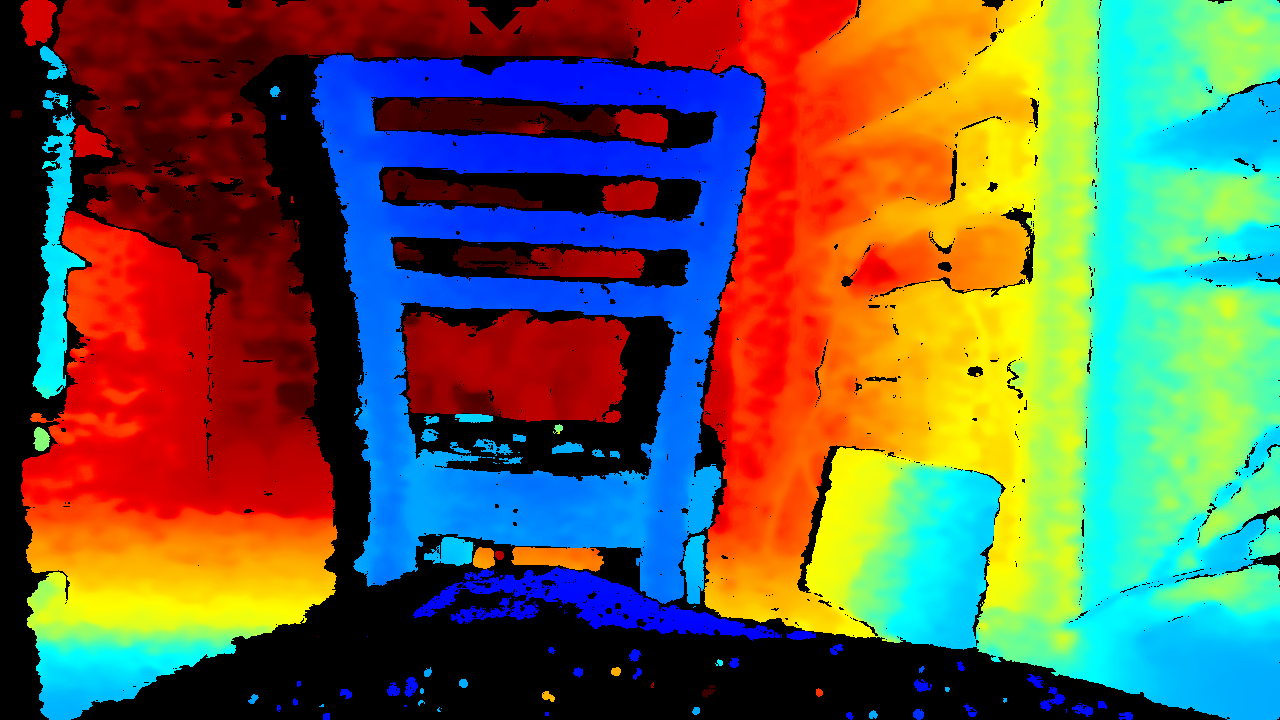} &
			\includegraphics[width=0.3\linewidth]{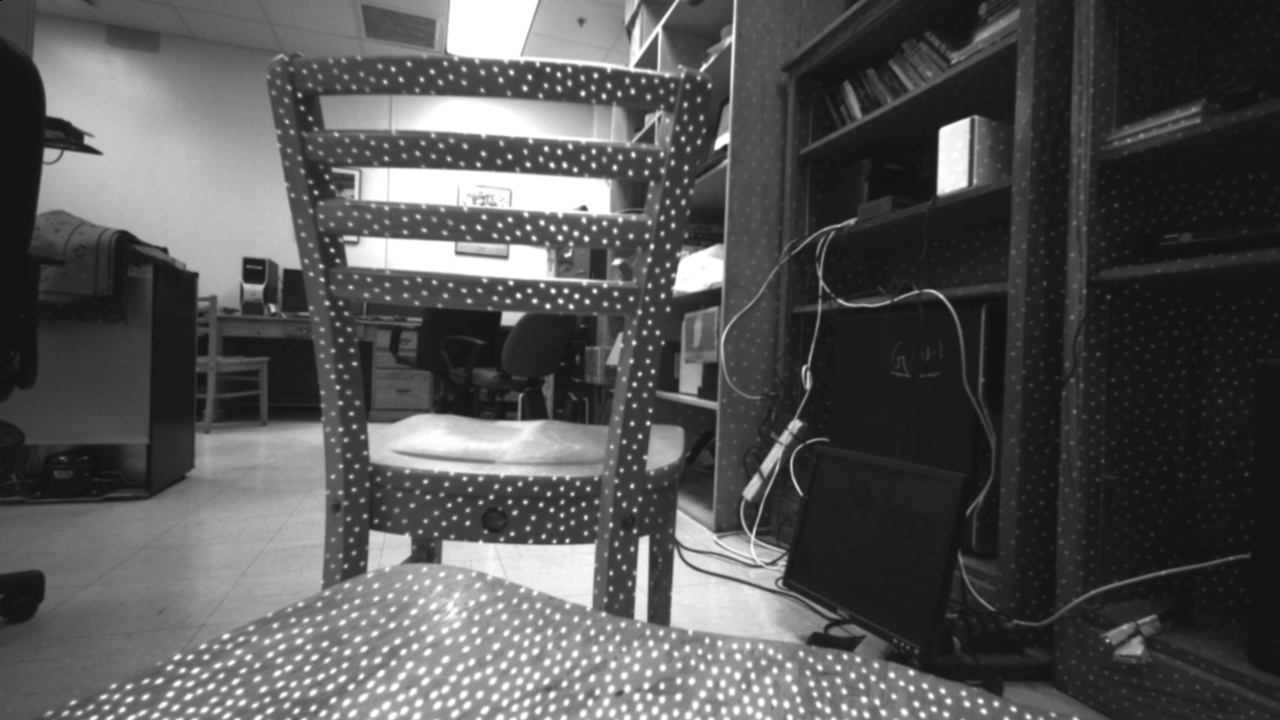} &
			\includegraphics[width=0.3\linewidth]{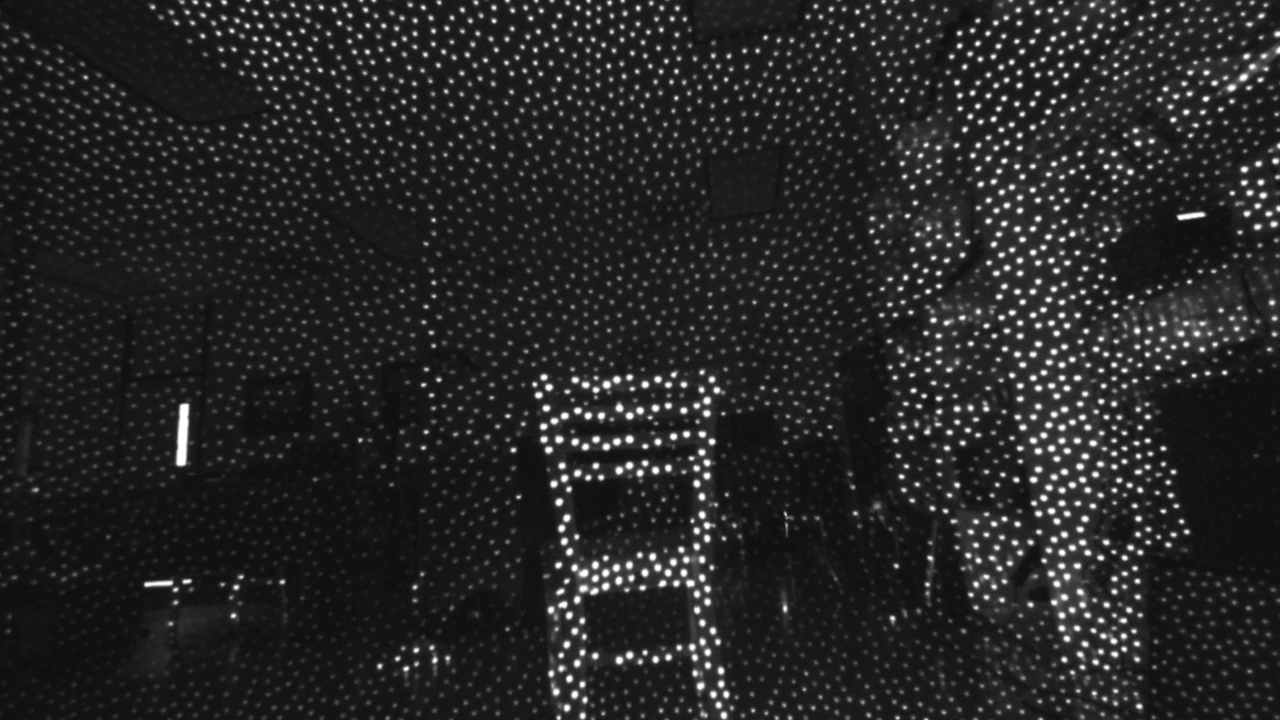} \\
			(a)& (b)& (c)
		\end{tabular}
		
	\end{center}
	\caption{Sample pictures obtained from depth camera.}
	\label{fig:depthinput}
\end{figure}

 In the Gazebo simulated training environment, the robot is equipped with the depth camera. From this depth camera, the robot can sense depth, color, and texture from the RGB sensor. Figure \ref{fig:depthinput} shows sample pictures from the depth camera. The blobs in the pictures are laser beams which form PC. Picture (a) is a depth image taken during the daytime, (b) is an infrared image also taken during the daytime, and (c) is an infrared image taken at night. The PC is the input to RL both on the real sidewalk as well as in simulation. The base moving speed of the robot is set to the mean walking speed of men to make the simulation close to the real sidewalk. The lighting condition is set to ambient light, which gives an illusion of daylight. We were also able to set the wind speed of the ambient environment. There were ways to create a sidewalk with a slippery surface, with ice, snow, and slope. However, to avoid the extreme complexity of the implementation, we skipped these aspects within this research scope.

In the training environment, we examined the learning of the three algorithms. The same sidewalk was used for training all of these. Q-learning, State-Action-Reward-State-Action (SARSA), and Deep Q-Network (DQN) belong to the list of training algorithms. The Q-learning algorithm is off-policy meaning it learns based on the action obtained from another policy e.g., greedy approach. Whereas the SARSA algorithm is on-policy that learn based on the action performed by the current policy instead of the greedy approach. Both Q-learning and SARSA are not generalized because these algorithms have to experience a state before learning. That is why these are not generalized and performs poorly in a huge number of states. 

\begin{figure}[h!]
	\begin{center}
		\includegraphics[width=\linewidth]{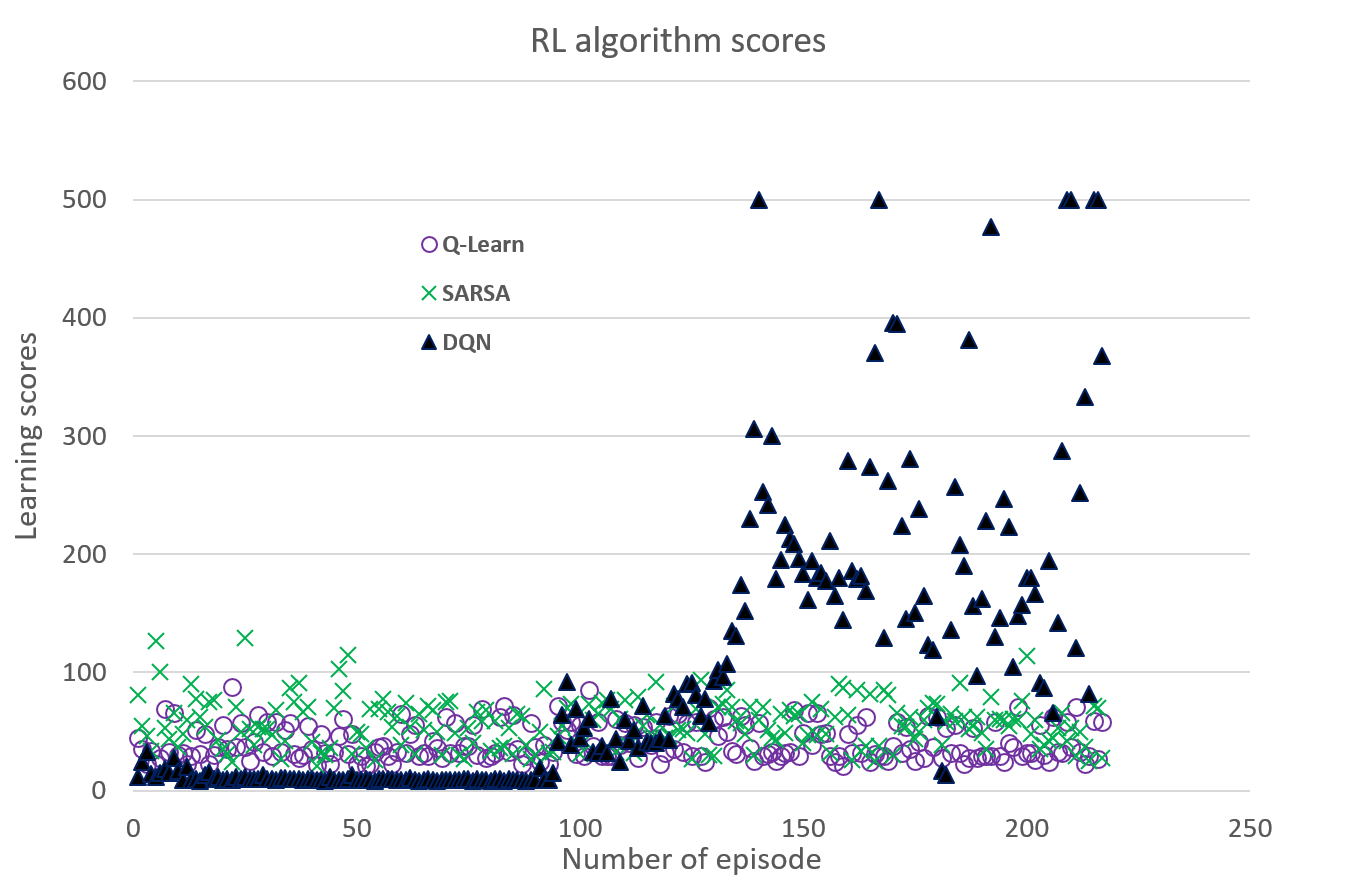}
	\end{center}
	\caption{Learning score comparison of Q-learning, SARSA, and DQN.}
	\label{fig:learningscores}
\end{figure}

Within 200 episodes, the DQN learned best among the three, and SARSA learned better than the Q-learning. Figure \ref{fig:learningscores} shows these findings. The derivative of the learning curve of the reward increased over the number of episodes. In other words, we can say that the more interaction the robot makes with the obstacles, the more it learns to avoid them. That is why the reward increases after a couple hundred episodes. 

The testing environment of the RL model is the real sidewalk. A visually impaired person volunteered to test the prototype. The IRB approval of the blind-ambition umbrella project is used for this testing as it involves human subjects. Five hundred feet of the u-shape sidewalk was selected for the evaluation of the prototype. There were trees, an electric pole, a pothole, a dumpster, an iron fence, a visible curb, a bollard, and a fire hydrant on this sidewalk. We manually placed a couple of electric scooters, yellow construction cones, and water to form a puddle. The user was mostly happy about being aware of upcoming objects ahead of time. He could quickly point to any direction and ask ``What is there?'' and receive names of the segmented objects. The obstacles which stand above the ground were found easily, but the ground level obstacles such as the pothole and puddle was hardly found. On a narrow sidewalk, the RL got confused with the sidewalk fence (not the construction fence) as an obstacle. Though there are limitations, according to the volunteer, the overall performance of the assistive device was found to be satisfactory. 
 
RASA \cite{bocklisch2017rasa} framework is the base engine for building a conversational agent. To train it,  it requires conversational data, which we have obtained from the WOZ experiment on the sidewalk. We carefully annotated the spoken sentences of the visually impaired into proper intent, and we identified the entities and actions from those. Executing actions requires developing a service engine. An entity is passed as a parameter to the action. The RASA stack provides a light-weight SDK for this purpose.  We used this SDK to develop the action end-point.  

The input and output of the conversational agent is text. From an audio input device, the speech is converted to text and fed into the agent. The reply from the agent was again converted back to speech and sent to the audio output device. We have used the speech-to-text engine for the speech to text conversion, and it generated words with correct spelling. Because the RASA stack always receives words with correct spelling, we did not have to train it with incorrectly spelled words. For example, we avoided training the conversational agent with the variation of ``hi'', ``hey'', or ``hai''. 

The Bluetooth headset acts as an audio input and output interface. This device connects to the prototype of the assistive device and provides a partial scope of the private conversation. That is, people may hear what the visually impaired person is asking for, but they cannot hear what the device is replying. 

We show the basic block diagram (see Figure \ref{fig:agtblockdiagram}) of the prototype. The user has the option to ask the AG to take a picture and segment it. Amazon Rekognition does the segmentation of images in AG. The text-to-speech and speech-to-text service is used from Google. Of course, to use the Google and AWS services, there is a need for internet connectivity.

Table \ref{tab:obstacleavoid} contains the results obtained from a test simulation. In the testing phase, we let the robot walk from one side to the other side of the sidewalk $10,000$ times which is the number of episodes. The robot found the construction cone most of the time but failed to see the pothole. It is reasonable, because the pothole is on the ground whereas the construction cone, fire hydrant, stopper, and electric scooter stand above the ground. Among the above ground level obstacles, the AG is less able to detect the electric scooter than other obstacles. This less detection is due to the size and shape of the scooter. 

\begin{figure}[h!]
	\begin{center}
           \begin{tabular}{ccc}
		\includegraphics[width=0.3\linewidth]{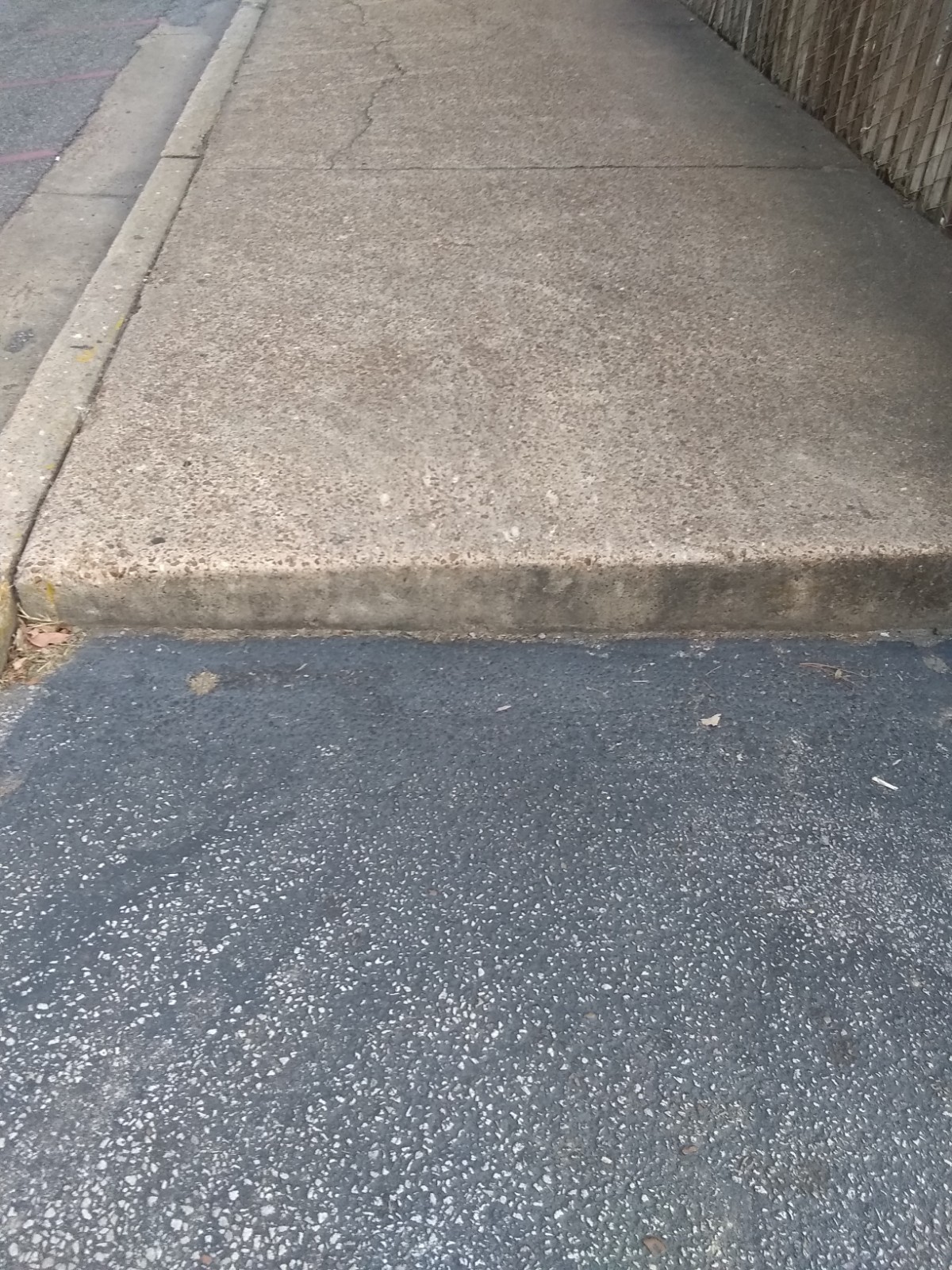} &
		\includegraphics[width=0.3\linewidth]{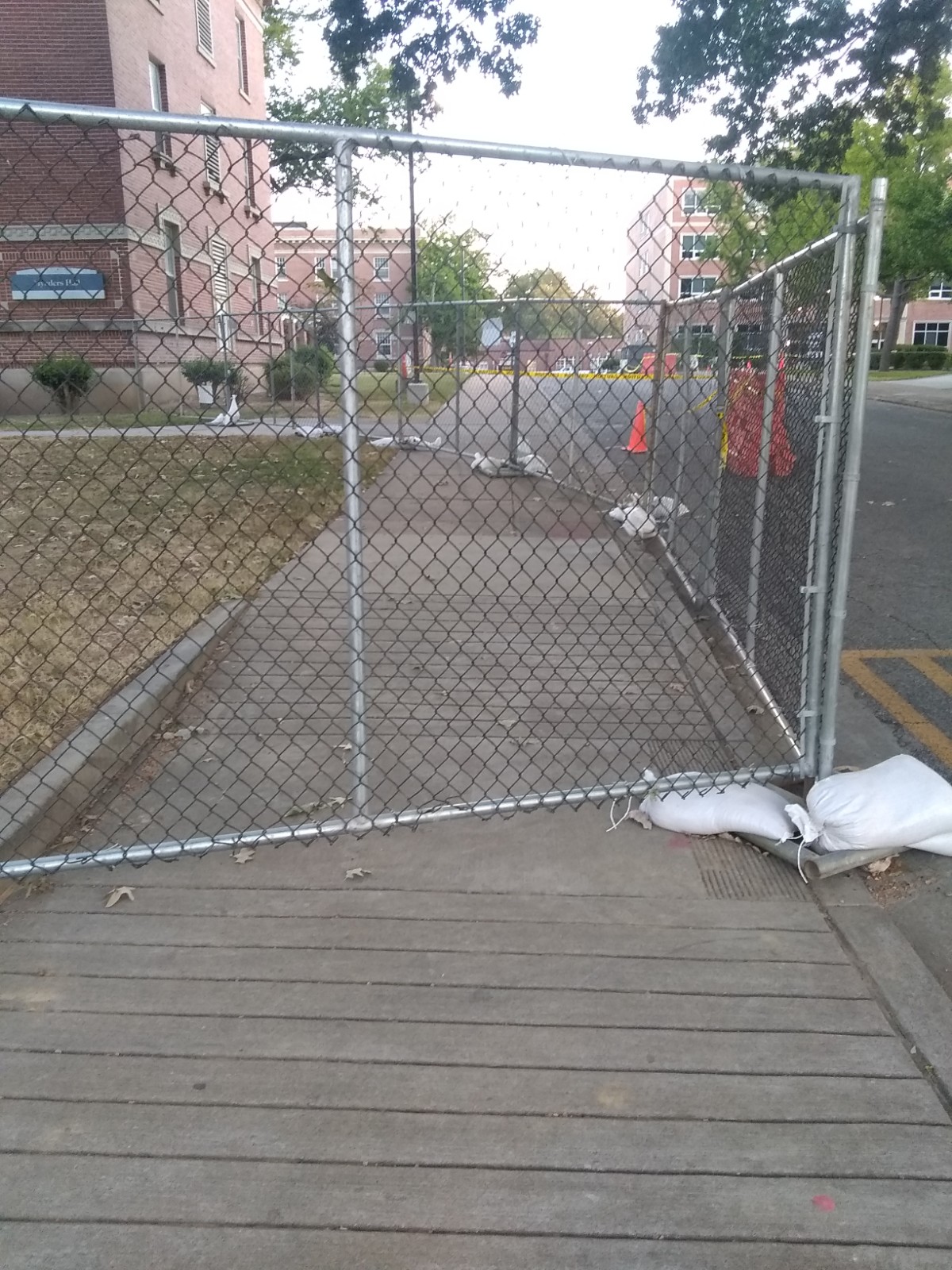} &
		\includegraphics[width=0.3\linewidth]{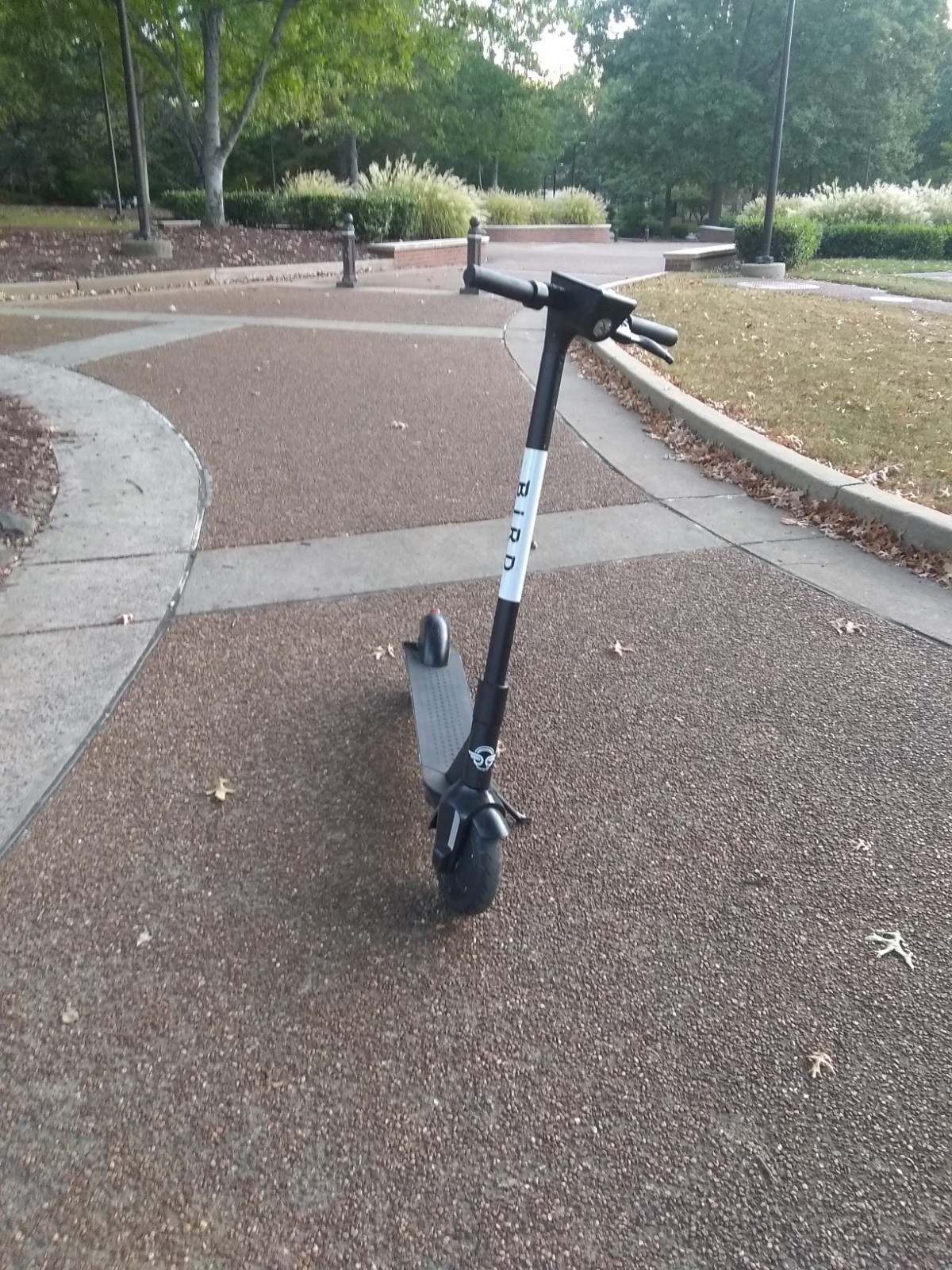} \\
		(a) carb & (b) fence & (c) scooter \\
		\includegraphics[width=0.3\linewidth]{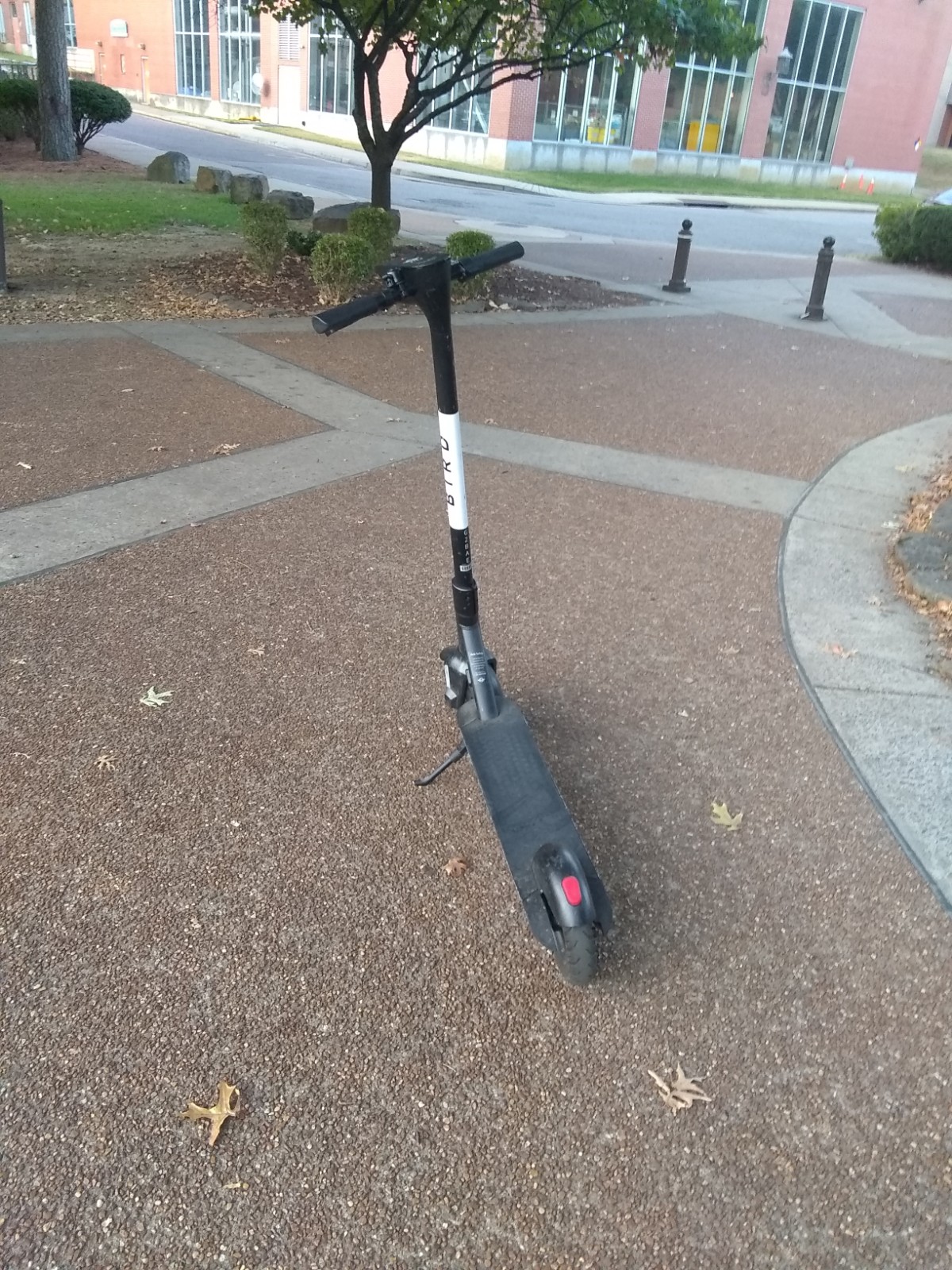} & 
		\includegraphics[width=0.3\linewidth]{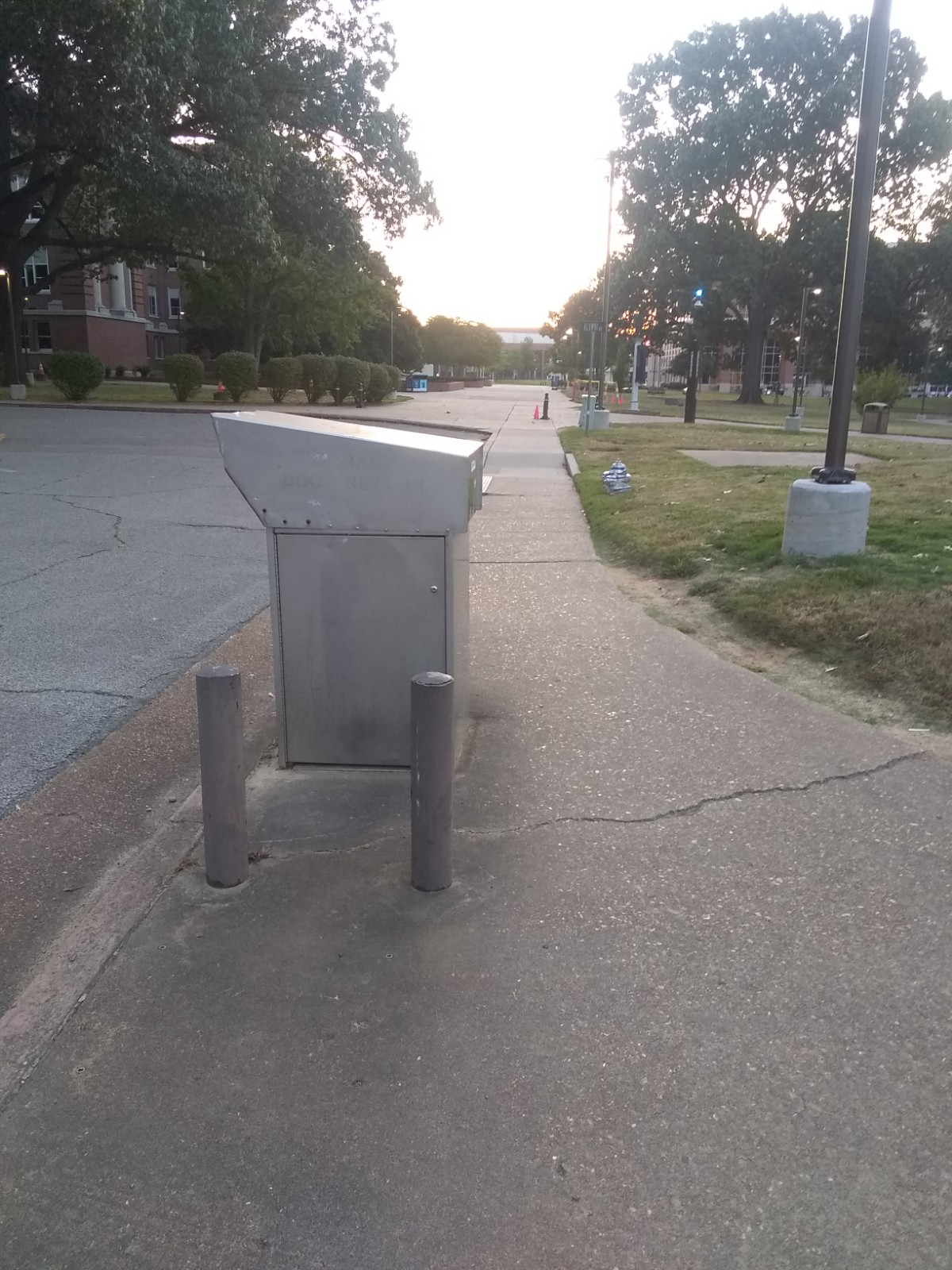} &
		\includegraphics[width=0.3\linewidth]{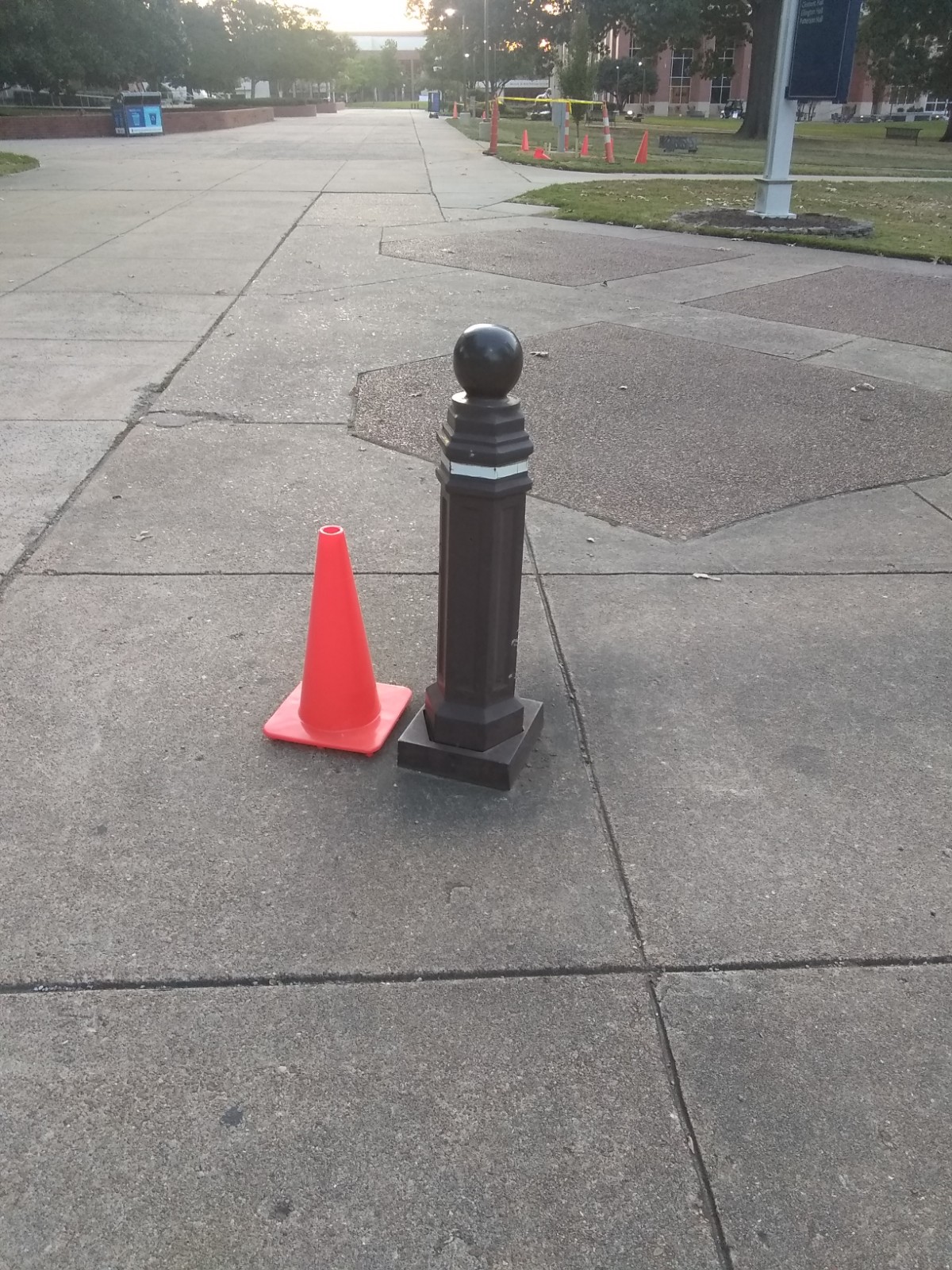} \\
                      (d) scooter & (e) dropbox & (f) stopper 	    
           \end{tabular}	
	\end{center}
	\caption{Example obstacles detected by the assistive device}\label{fig:otherobstacles}
\end{figure}

For comparison, we selected the base case as 78.75\% ``in field obstacle detection accuracy'' from \cite{ahmed2018image}. We used the AG on the real sidewalk and found that the average accuracy measured is about 81.29\%. The obstacle avoidance experience improved about 2.5\%.  It detected and talked about various obstacles. Few important obstacles are shown in Figure \ref{fig:otherobstacles}.

\begin{table}[h!]
	\begin{center}
		\caption{Obstacles avoidance results in simulation and real world}\label{tab:obstacleavoid}
		\begin{tabular}{m{0.1\textwidth}cm{0.1\textwidth}m{0.1\textwidth}}
			\hline 
			Obstacle & Image &\% simulation & \% sidewalk\\ 
			\hline 
			Pothole & \includegraphics[width=0.2\linewidth,height=5em]{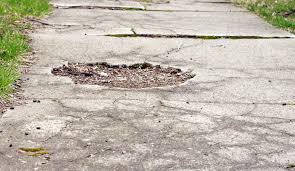} & 55 & 52\\ 
			Construction Cone& \includegraphics[width=0.2\linewidth,height=4em]{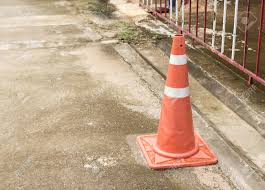} & 91 & 92 \\ 
			Fire hydrant & \includegraphics[width=0.2\linewidth,height=4em]{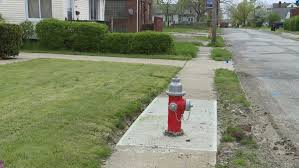} & 92 & 93\\ 
			Electric Scooter &\includegraphics[width=0.2\linewidth,height=4em]{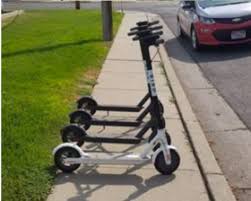} & 74 & 71\\ 
			Electric Pole & \includegraphics[width=0.2\linewidth,height=4em]{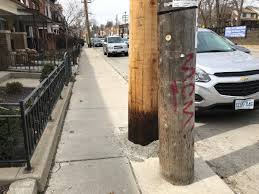} & 78 & 79\\ 
			Dumpster & \includegraphics[width=0.2\linewidth,height=4em]{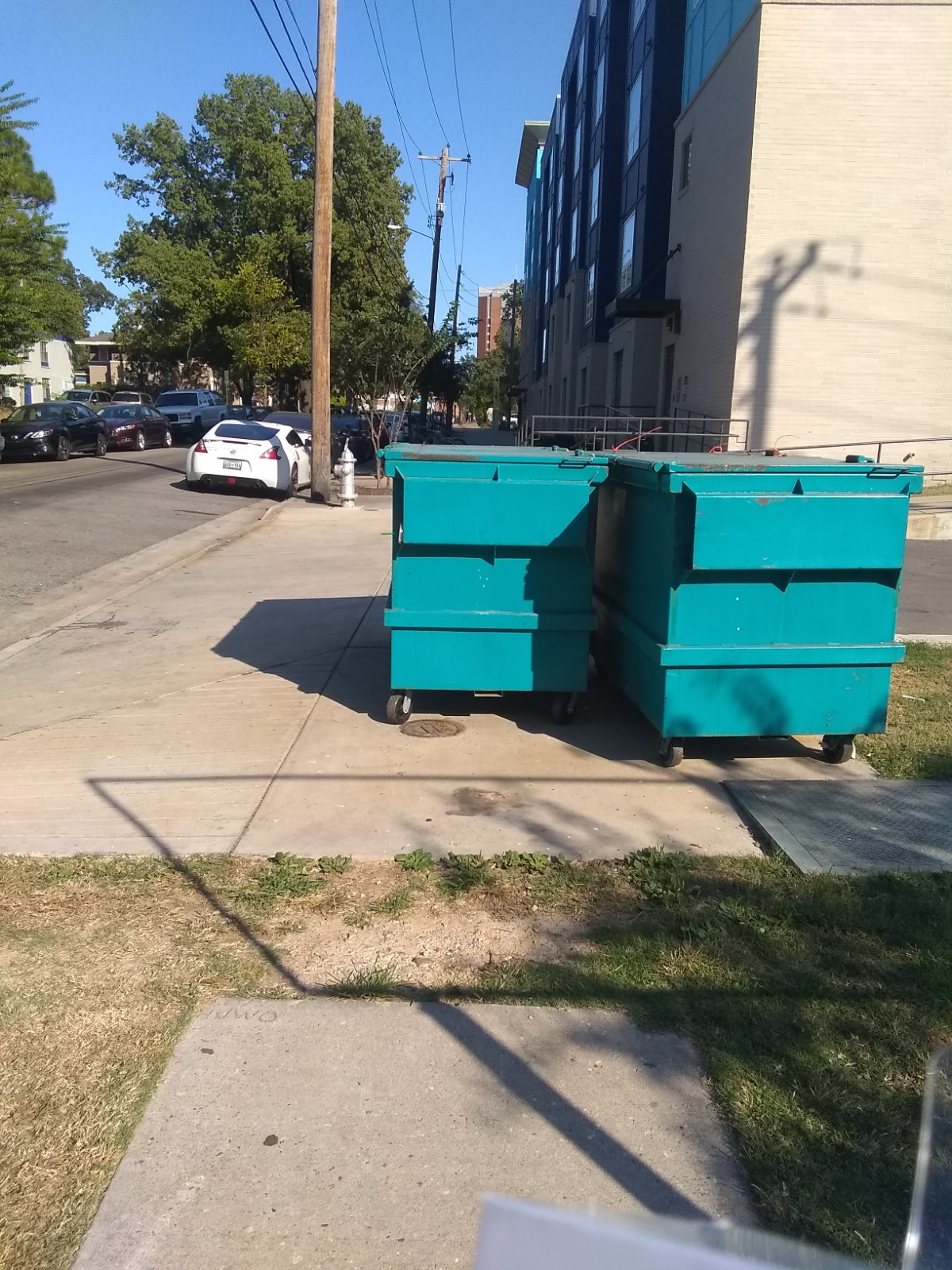} &  93 & 94\\ 
			Tree & \includegraphics[width=0.2\linewidth,height=4em]{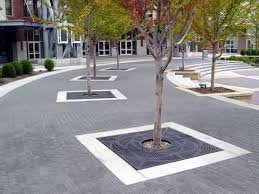}  & 87 & 89\\ 
			\hline 
			\hline 
			 & average & 81.428 & 81.285 \\
			\hline  
		\end{tabular} 
		
	\end{center}
\end{table}

\begin{figure}[h!]
	\begin{center}
		\includegraphics[width=0.5\linewidth]{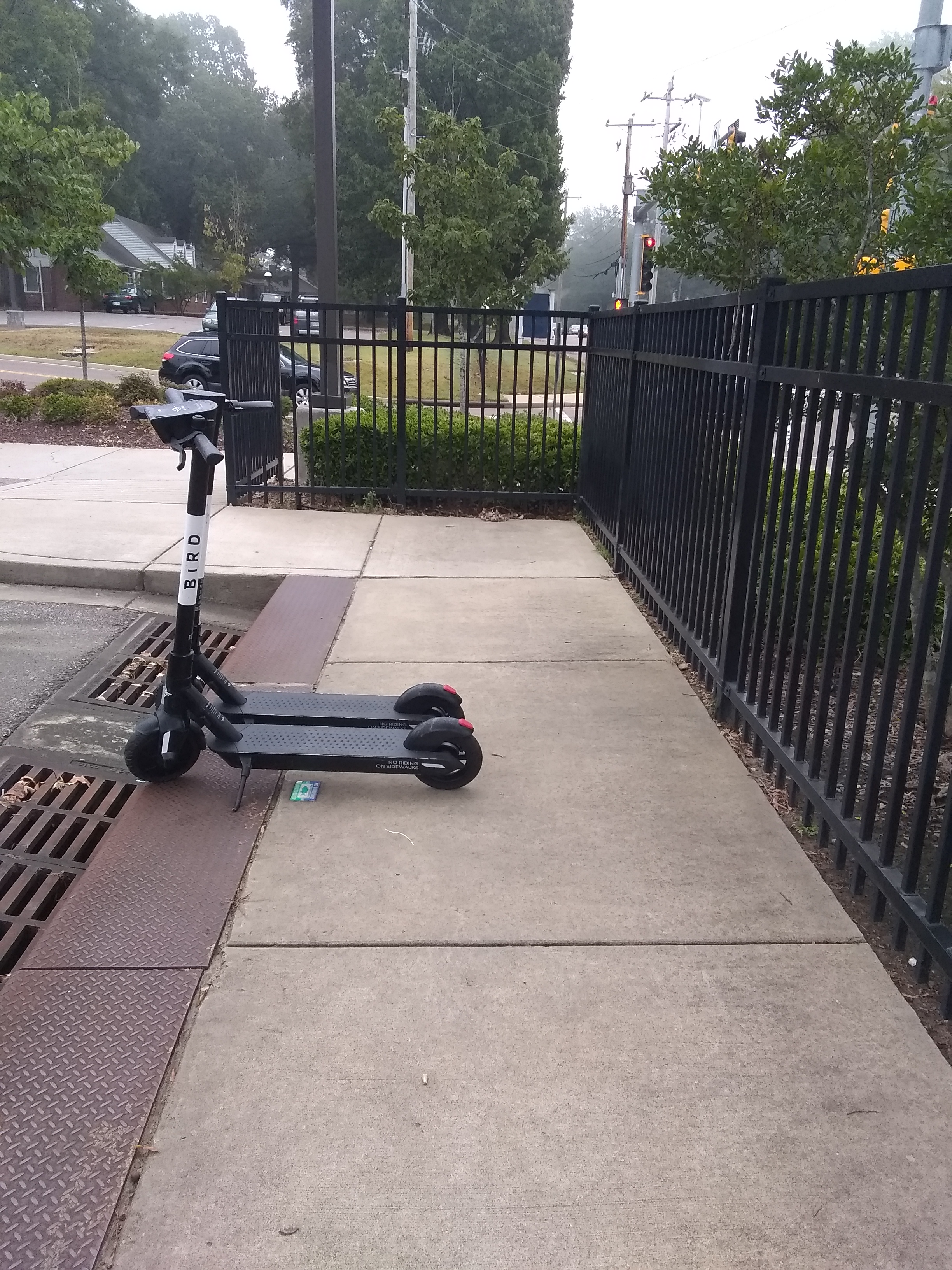} 	
	\end{center}
	\caption{AG got confused where scooter and sidewalk fence were together. It reported fence as obstacle instead of scooter.}\label{fig:confusedobstacle}
\end{figure}

The AG got confused with the obstacle in Figure \ref{fig:confusedobstacle}. The real obstacle is the electric scooter but it was talking about the fence as an obstacle as well. 

\subsection{Usage analytic}
The RL model can be evaluated in laboratory but how to evaluate the system. One way to evaluate system to give it to the user and they describe pros and cons. Another way is using Likert scale 
or Nasa TLX which is subjective. The Likert scale \cite{} and Nasa TLX \cite{} gives good rating which is sometimes biased for a system because users tends to say good about the system in front of the builder. To evaluate using Likert scale and NASA TLX we need to either bring the user in the lab or we go to them to ask them questions about the system.

The evaluation system should be easier for the user, they don’t have to come to the lab. They can evaluation at their convenient time and we can keep track of it. Moreover, due to any pandemic situation finding users to evaluate a system is harder. That is why we adopted a method of ``usage analytic''. It refers to collect, analyze, and visualize data on the use of an application. Through usage analytic we can understand user interactions and engagement along with system behavior. 

Augmented Guide evaluation require two parts of evaluation one is the software application itself another is how well it recognize obstacles. In addition how frequently the application crashes or how long it takes to respond also important. We found the following items should be included in usage analytic-
\begin{itemize}
    \item Text of the conversation
    \item Recognized obstacle
    \item Unrecognized obstacle
    \item Log of the application
    \item Keep alive log 
\end{itemize}

\begin{figure}[h!]
	\begin{center}
		\includegraphics[width=0.9\linewidth]{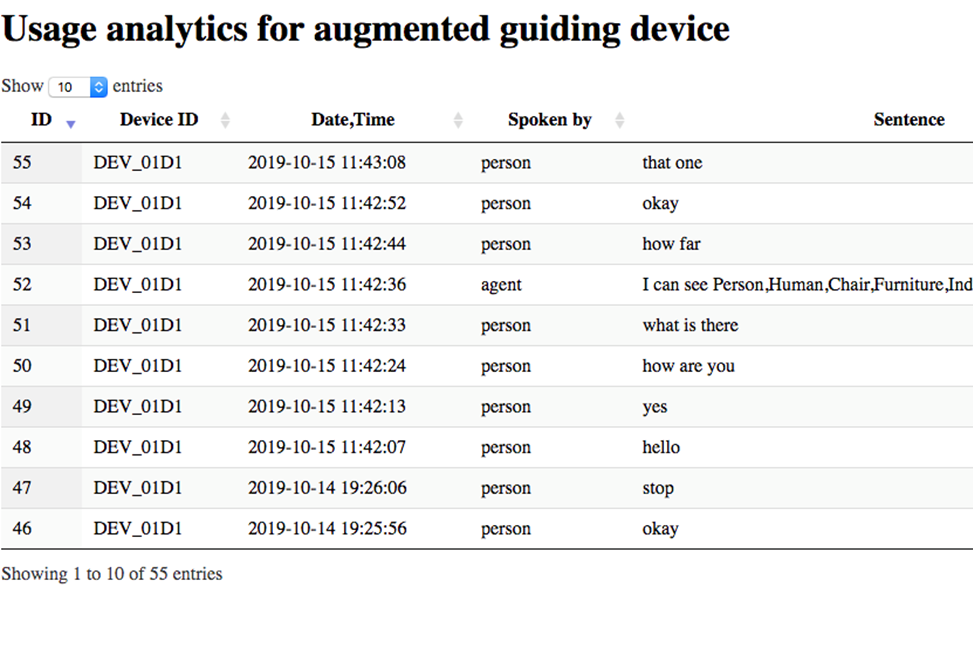} 	
	\end{center}
	\caption{Usage analytics screenshot.}\label{fig:usageanalytics}
\end{figure}

A screenshot of usage analytics is give in Fig. \ref{fig:usageanalytics}. The system deadlock time, System offline time, Frequency of usage, Task completion time, and User try are inferred usage analytics.

\section{Conclusion}
In this research, we built an assistive technology prototype device. The purpose of this prototype is to augment the means of avoiding obstacles for the visually impaired.  The obstacles include both static and dynamic nature, and the device is useful during a walk on the sidewalk. 



We developed the free path approach instead of modeling the various obstacles. Reinforcement learning served as an essential tool for free path modeling. Also, to communicate the free path to the user, we incorporated the conversational agent trained on the RASA stack. 

For modeling the free path, we created the simulated sidewalk and the 3D models of obstacles in Gazebo. We placed a robot in the environment, which learned to avoid obstacles through RL. When the RL was stable and produced SOAA model, we incorporated it into the AG. 

The conversational agent SOCA is trained with the Wizard of OZ conversation data. This conversation is the starting point of the agent learning to talk. The user asks the agent about the ambient environment. The agent then talks back to the user with the necessary information. RASA collects that information from AWS, API, and the RL model. From this information about the ambient environment, the user decides to take necessary actions.

We observed some limitations of the assistive prototype system during the training and testing. One of those is that the Gazebo obstacles are a purely mathematical model. It means that the physics engine sees a tree as a box even though the tree has a particular shape. During testing, we found that this limitation did not matter much because the input to the RL was PC. The conversation tool sometimes takes a long time to respond. It could be dangerous in a situation where time is crucial, e.g., an oncoming car while crossing the road. The use of the WiFi network is another limitation. It could be solved by keeping the models and services all in the computing device, but that requires higher computing, storage, and battery capacity. As a trade-off, the WiFi is used. Also, the most critical obstacle, according to Ahmed et al., is the ``slope''  \cite{ahmed2017optimization}. Our assistive device can not detect slope.

\bibliographystyle{IEEEtran}
\bibliography{main}

\end{document}